\documentclass{article}

\usepackage{PRIMEarxiv}

\usepackage[utf8]{inputenc} 
\usepackage[T1]{fontenc}    
\usepackage{hyperref}       
\usepackage{adjustbox}
\hypersetup{
    colorlinks=true,
    linkcolor=blue,
    filecolor=cyan,      
    urlcolor=magenta,
    }
\usepackage{url}            
\usepackage{booktabs}       
\usepackage{framed,multirow}
\usepackage{amsfonts,amssymb,amsmath}       
\usepackage{nicefrac}       
\usepackage{microtype}      
\usepackage{lipsum}
\usepackage{fancyhdr}       
\usepackage{graphicx}       
\usepackage{float}
\usepackage{wrapfig}
\usepackage{xcolor}
\usepackage{soul}
\usepackage[nameinlink]{cleveref}
\graphicspath{{Figures/}}     
\usepackage{subcaption}
\usepackage{caption}
\usepackage{vcell}
\usepackage{colortbl}
\usepackage{makecell}

\usepackage{romannum}
\AtBeginDocument{\pagenumbering{arabic}}

\title{Loss Functions in the Era of Semantic Segmentation: A Survey and Outlook}

\author{
	Reza Azad* \\
	Faculty of Electrical Engineering \\ 
 and Information Technology,\\
	RWTH Aachen University\\
	Aachen, Germany\\ \And
	Moein Heidari* \\
	School of Electrical Engineering \\
	Iran University of Science and Technology \\
	Tehran, Iran\\ \And
	Kadir Yilmaz* \\
	Faculty of Electrical Engineering \\ 
 and Information Technology,\\
	RWTH Aachen University \\
	Aachen, Germany \\ \And
	Michael Hüttemann* \\
	Faculty of Electrical Engineering \\ 
 and Information Technology,\\ 
	RWTH Aachen University \\
	Aachen, Germany \\ \And
	Sanaz  Karimijafarbigloo* \\
	Faculty of Informatics and Data Science\\ 
	University of Regensburg \\
	Regensburg, Germany \\ \And
	Yuli Wu \\
	Faculty of Electrical Engineering \\ 
 and Information Technology,\\
	RWTH Aachen University \\
	Aachen, Germany \\ \And
	Anke Schmeink \\
	Faculty of Electrical Engineering \\ 
 and Information Technology,\\
	RWTH Aachen University \\
	Aachen, Germany \\ \And
        Dorit Merhof \\
        Faculty of Informatics and Data Science\\ 
    	University of Regensburg \\
    	Regensburg, Germany \\ 
        \texttt{dorit.merhof@ur.de} \\
}

\begin{document}
\maketitle
\begin{abstract}
Semantic image segmentation, the process of classifying each pixel in an image into a particular class, plays an important role in many visual understanding systems.
As the predominant criterion for evaluating the performance of statistical models, loss functions are crucial for shaping the development of deep learning-based segmentation algorithms and improving their overall performance.
To aid researchers in identifying the optimal loss function for their particular application, this survey provides a comprehensive and unified review of $25$ loss functions utilized in image segmentation.
We provide a novel taxonomy and thorough review of how these loss functions are customized and leveraged in image segmentation, with a systematic categorization emphasizing their significant features and applications.
Furthermore, to evaluate the efficacy of these methods in real-world scenarios, we propose unbiased evaluations of some distinct and renowned loss functions on established medical and natural image datasets. We conclude this review by identifying current challenges and unveiling future research opportunities.
Finally, we have compiled the reviewed studies that have open-source implementations at our \href{https://github.com/YilmazKadir/Segmentation_Losses}{GitHub}\footnote{\url{https://github.com/YilmazKadir/Segmentation_Losses}\\ {* Equal contribution}}.
\end{abstract}


\section{Introduction}
Image segmentation plays a fundamental role in a wide range of visual understanding systems. Its primary objective is to generate a dense prediction for a given image, $i.e.$, assigning each pixel a pre-defined class label (semantic segmentation)~\cite{krahenbuhl2011efficient,azad2023foundational}, or associating each pixel with an object instance (instance segmentation)~\cite{hariharan2014simultaneous}, or the mixture of both (panoptic segmentation)~\cite{Kirillov2018PanopticS}, which facilitates the organization of pixels with similar semantics into meaningful high-level concepts. Segmentation has a wide range of applications in a variety of domains including medical image analysis~\cite{heidari2023hiformer,azad2023unlocking}, video surveillance, and augmented reality~\cite{9356353}, to name a few.
From Convolutional Neural Networks (CNN) up to Transformers, many diverse model architectures have been introduced for semantic segmentation~\cite{azad2022medical,azad2023advances}. However, the optimal performance of segmentation models depends on the choice of the right network structure and the appropriate objective function. Specifically, a prominent field of study in image segmentation involves developing approaches to alleviate diverse challenges including class imbalance, scarce availability of datasets and noisy, human-biased, and poor inter-annotator agreement labels by widely proposing robust loss functions to allow joint optimization of model parameters~\cite{gonzalez2023robust}. Furthermore, numerous modern deep image segmentation techniques are prone to fail in recovering thin connections, intricate structural elements, precise boundary positioning, and subsequently, the correct topology of images~\cite{ngoc2021introducing}. Due to the surge of interest from the research community to mitigate these problems, a survey of the existing literature is beneficial for the community and timely to help avid researchers and practitioners exploit the best objective function for their segmentation task at hand.
Specifically, this review provides a holistic overview of $25$ loss functions developed for image segmentation applications. We provide a taxonomy of their design, highlight the major strengths and shortcomings of the existing approaches, and review several key technologies arising from various applications, including natural and medical image segmentation. We provide comparative experiments of some reviewed methods on two popular medical and natural image datasets and offer their codes and pre-trained weights on \href{https://github.com/YilmazKadir/Segmentation_Losses}{GitHub}. 
A thorough search of the relevant literature revealed that we are the first to cover the loss functions used in the semantic segmentation domain following~\cite{Jadon2020ASO}. However, in contrast to~\cite{Jadon2020ASO}, we propose a new detailed and organized taxonomy, highlight task-specific challenges, and provide insights on how to solve them based on the reviewed literature, which allows a structured understanding of the research progress and limitations in different areas by covering research works beyond $2020$.

\begin{figure*}[t]
 \centering
 \includegraphics[width=1\textwidth]{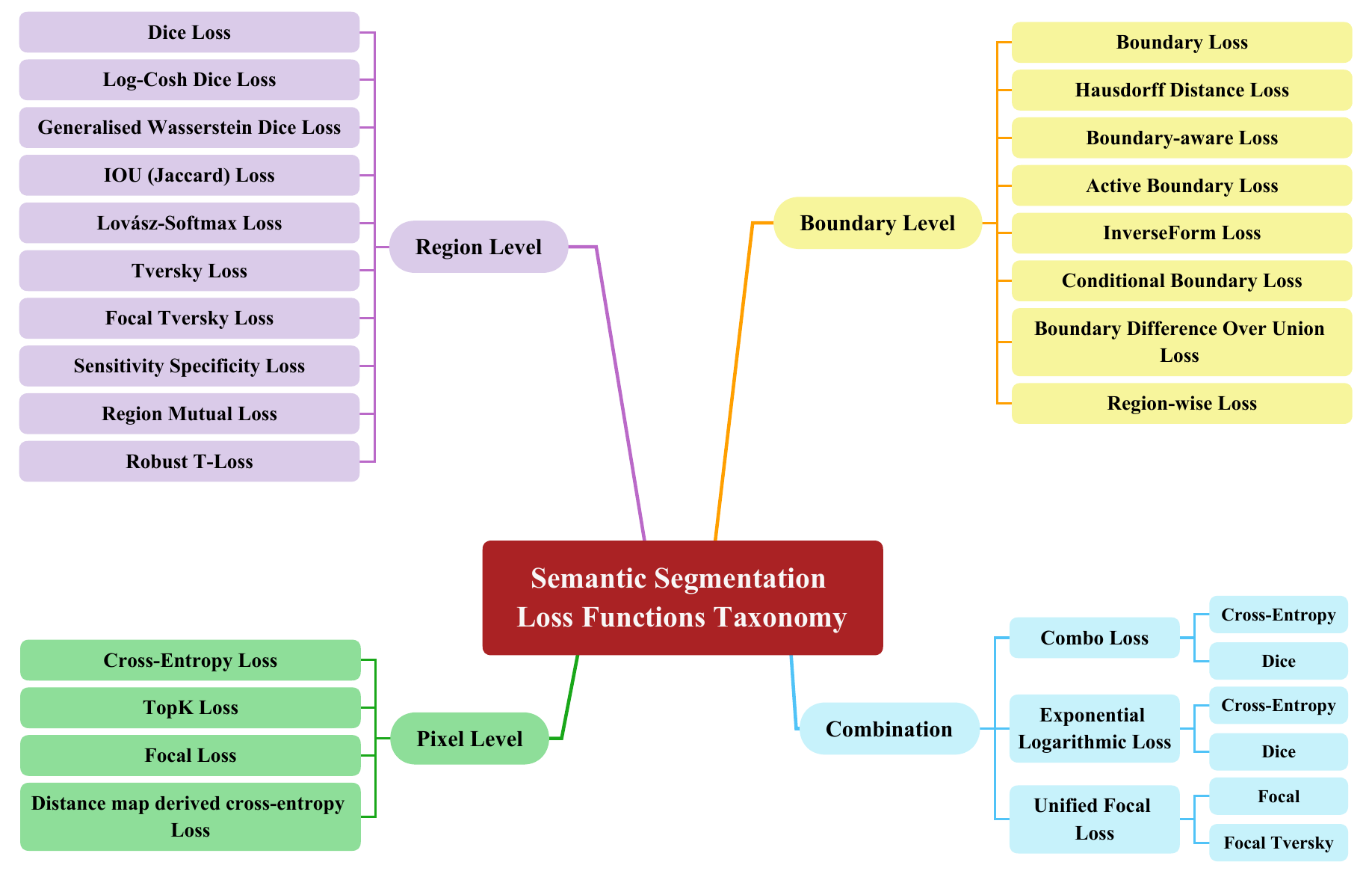}
 \caption{The taxonomy subsections delineate four distinct groups: (1) Region-Level, (2) Boundary-Level, (3) Pixel-Level, and (4) Combination.}
 \label{fig:taxonomy}
 \vspace{-1em}
\end{figure*}

Additionally, we present extensive qualitative and quantitative experiments that validate design decisions and performance on widely recognized datasets for each taxonomy in both natural and medical image segmentation. 
Furthermore, when considering how loss functions affect both CNN- and Transformer-based approaches. We believe that this work will highlight new research opportunities, provide guidance to researchers, and stimulate further interest in the Computer Vision community to leverage the potential of covered loss functions in the segmentation domain.
Some of the key contributions of this review paper can be outlined as follows:

\begin{itemize} 

\item We systematically and thoroughly examine the loss functions in the field of image segmentation and provide a comparison and analysis of these approaches. Specifically, $25$ loss functions in semantic segmentation are covered in a hierarchical and structured manner.

\item Our work provides a taxonomized (\Cref{fig:taxonomy}), in-depth analysis of the loss function, as well as a discussion of various aspects.

\item We perform comparative experiments on a selection of the methods we've reviewed, using two well-known datasets for segmenting both natural and medical images: Cityscapes~\cite{7780719} and the Synapse multi-organ segmentation~\cite{synapse2015ct} dataset.

\item Finally, we address obstacles and unresolved issues, while also acknowledging emerging patterns, presenting unanswered questions, and identifying possible directions for future investigation.

\end{itemize}

\subsection{Motivation and uniqueness of this survey}
Image segmentation approaches have undergone significant advances over the past few decades. These range from the prevailing direction of integrating multi-resolution and hierarchical feature maps~\cite{heidari2023hiformer}, to the exploitation of boundary information~\cite{xiao2023baseg,10.1007/978-3-031-21014-3_39}, and to the joint optimization of semantic segmentation and supplemental tasks in a multitask learning framework~\cite{wang2022semi}. These diverse methods are advocated to mitigate certain challenges in segmentation including class imbalance, erroneous or incomplete boundaries, and pixel importance~\cite{VALVERDE2023109208}, to name a few. Numerous strategies have attempted to alleviate these issues by proposing improved loss functions.
Although one survey paper was already been published before this area was fully developed\cite{jadon2020survey}, much progress has been made in this area since then. On the other hand, no survey paper focuses on an application-oriented perspective on loss functions in segmentation, which is the central aspect in pushing this research direction forward. Thus, there is a clear gap in the community. 
More importantly, loss functions serve as critical tools in the process of training machine learning models to accurately delineate regions of interest within images. In the medical domain, accurate segmentation can have life-saving implications by aiding in the diagnosis and treatment of disease. Different loss functions can greatly affect the model's ability to segment anatomical structures or detect abnormalities within medical images. Similarly, in the natural image domain, such as satellite imagery or scene understanding, accurate segmentation is critical for applications such as autonomous vehicles and environmental monitoring. The choice of a loss function can also significantly affect the model's performance in these domains. 
Hence, in our survey, we guide readers from both the medical and computer vision communities to understand the objective and use cases of these loss functions. 
Additionally, by evaluating these loss functions in the context of both Convolutional Neural Networks (CNN) and Transformer-based approaches, across diverse domains encompassing medical and natural images, we aim to illustrate their true efficacy in addressing challenging tasks. This comprehensive examination of loss functions is expected to provide readers with a broader perspective for making informed decisions regarding the adoption of more appropriate loss functions.

\subsection{Search Strategy}

We conducted searches on platforms such as DBLP, Google Scholar, and Arxiv Sanity Preserver, taking advantage of their ability to generate customized search queries and comprehensive lists of academic works. These searches encompassed a wide range of scholarly publications, including peer-reviewed journal articles, papers presented at conferences or workshops, non-peer-reviewed materials, and preprints, all achieved through customized search criteria. Our specific search query was \texttt{(loss* deep $|$ segmentation*) (loss $|$ segmentation*) (loss* $|$ train* $|$ segmentation* $|$ model*) (loss* $|$ function* $|$ segmentation* $|$ medical*)}. We filtered our search results to remove false positives and included only papers related to semantic segmentation models. In the end, we chose to delve deeper into the different loss functions that are commonly used in the existing literature or that are designed for a particular purpose.

\subsection{Paper Organization}

The remaining sections of the paper are organized as
follows. In Section~\ref{sec:loss-in-action}, we provide a detailed overview of the key components of the well-established loss function in image segmentation. Moreover, this section clarifies the categorization of objective function variants by proposing a taxonomy to characterize technical innovations and important use cases. For each loss function, we present the theoretical foundation and fundamental concepts as well as open challenges and future perspectives of the field as a whole.
In Section~\ref{sec:experiment}, we evaluate the performance of several previously discussed loss function variants on favoured natural/medical segmentation benchmarks.
Finally, Section~\ref{sec:conclusion} summarizes and concludes this review.

\section{Loss Functions in Semantic Segmentation}
\label{sec:loss-in-action}

We categorize existing work on semantic segmentation loss functions into three main groups based on their focus and objectives (see \Cref{fig:taxonomy}). 
Pixel-level loss functions operate at the individual pixel level and aim to ensure accurate classification of each pixel within the segmented regions. These loss functions calculate the discrepancy between the predicted pixel values and the corresponding ground truth labels independently for each pixel. 
In contrast, region-level loss functions focus on the overall class segmentation by maximizing the alignment between the predicted segmentation mask and the ground truth mask. These methods emphasize overlap and prioritize accurate object segmentation over pixel-wise details. Finally, boundary level loss functions specifically address the precision of object boundaries in the segmentation task to effectively separate overlapping objects. These losses work to minimize the distance or dissimilarity between the predicted boundary and the ground truth boundary, thus promoting fine-grained alignment of segmented regions. By categorizing loss functions into these three levels, i.e., pixel, region, and boundary, the field gains a comprehensive perspective on the various strategies employed to improve semantic segmentation performance.
To maintain consistency throughout the paper, we establish the formal notation, as depicted in~\Cref{tab:notation}, before elaborating on individual loss functions. All the formulations in this paper will adhere to this notation unless otherwise noted.

\begin{table}[t]
    \centering
    \caption{Notation for Segmentation Loss}
    \begin{adjustbox}{width=\columnwidth}
    \begin{tabular}{c|p{0.8\linewidth}}
        \rowcolor{gray!30}
        \textbf{\textcolor{black}{Symbol}} & \textbf{\textcolor{black}{Description}} \\
        \hline
        \rowcolor{blue!10}
        $N$ & Number of pixels \\
        \rowcolor{cyan!10}
        $C$ & Number of target classes \\
        \rowcolor{green!10}
        $t_n$ & One-hot encoding vector representing the target class of the $n^{\text{th}}$ pixel. \\
        \rowcolor{yellow!10}
        $t_n^c$ & Binary indicator: 1 if the $n^{\text{th}}$ pixel belongs to class $c$, otherwise 0. \\
        \rowcolor{orange!10}
        $y_n$ & Predicted class probabilities for $n^{\text{th}}$ pixel. \\
        \rowcolor{red!10}
        $y_n^c$ & Predicted probability of $n^{\text{th}}$ pixel belonging to class $c$. \\
        \rowcolor{purple!10}
        $t_n \cdot y_n$ & Predicted probability for the target class of $n^{\text{th}}$ pixel. \\
        \rowcolor{pink!10}
        $w$ & Weights assigned to target classes. \\
    \end{tabular}
    \end{adjustbox}
    \label{tab:notation}
\end{table}

In the following subsections, we will elaborate on each category in more detail.
\subsection{Pixel Level}
Pixel-level loss functions in semantic segmentation dive deep into the individual pixels to achieve high accuracy in classifying each pixel within segmented regions. These loss functions compute the dissimilarity or error between the predicted pixel values and their corresponding ground truth labels independently for each pixel. They excel in scenarios where fine-grained pixel-wise accuracy is paramount, such as in tasks requiring detailed object recognition and segmentation. Below, we present several well-known loss functions in this direction.

\subsubsection{Cross-Entropy Loss} Cross-entropy (CE) measures the difference between two probability distributions for a given random variable.
In segmentation tasks, the cross-entropy loss is used~\cite{Kline2005RevisitingSA} to measure how well the model's predictions match the target labels.
Using the softmax function, the model generates pixel-wise probability maps representing the likelihood of each pixel belonging to each class.
The cross-entropy loss is then calculated by taking the negative logarithm of the predicted probability for the target class at each pixel.
The cross-entropy loss approaches 0 as the predicted probability for the target class approaches 1.
\begin{equation}
L_{CE}(y, t)=-\sum_{n=1}^{N} \log (t_n \cdot y_n)
\label{eqn:CE}
\end{equation}
As $t_n$ is a one-hot encoded vector, only the predicted probability of the target class affects the cross-entropy loss.

When dealing with imbalanced datasets, one approach to cross-entropy loss is to assign different weights to each class.
This can help to balance the influence of each class on the overall loss and improve the performance of the model on the underrepresented classes.
One way to assign weights is to use the inverse class frequency, which means that the weight for each class is inversely proportional to the number of samples in that class.
So, classes with fewer samples will have higher weights and classes with more samples will have lower weights.

\begin{equation}
L_{WCE}(y, t, w) = -\sum_{n=1}^{N} t_n \cdot w \log (t_n \cdot y_n)
\label{eqn:WCE}
\end{equation}

For each pixel, the weight of the target class is used.
If all weights are set to 1, we simply get the cross-entropy loss.

\subsubsection{TopK Loss}
TopK loss \cite{wu2016bridging} is an extension of cross-entropy loss that causes the model to learn from only the hardest pixels to classify at each iteration.
The top k\% of pixels with the lowest predicted probability for the target class are selected, and only their loss is considered.
It can be formulated as:
\begin{equation}
L_{TopK}(y, t)=-\sum_{n \in K} \log (t_n \cdot y_n)
\label{eqn:TopK}
\end{equation}
where K is the set containing the \%k of pixels with the lowest probability assigned to the target class.

\subsubsection{Focal Loss} Another approach to deal with data imbalance is to use Focal loss~\cite{Lin2017FocalLF}.
Focal loss is a modified version of the cross-entropy loss that assigns different weights to easy and hard samples.
Here, hard samples are samples that are misclassified with a high probability while easy samples are those that are correctly classified with a high probability.
This helps to balance the influence of easy and hard samples on the overall loss.

\begin{equation}
L_{focal}(y, t, \gamma) = -\sum_{n=1}^{N} (1 - t_n \cdot y_n)^\gamma \log (t_n \cdot y_n)
\label{eqn:Focal}
\end{equation}
where $\gamma$ is a non-negative tunable hyperparameter.
When $\gamma$ is set to 0 for all samples, we get plain cross-entropy loss.

\subsubsection{Distance Map Derived Cross-Entropy Loss} Many semantic segmentation models suffer from performance degradation at object boundaries.
To make the model focus on hard-to-segment boundary regions, a straightforward approach is to penalize segmentation errors at object boundaries harder.
To this end, Caliva et al.~\cite{Caliv2019DistanceML} uses distance maps.
A distance map has the same shape as the image and each pixel is assigned as the pixel's smallest distance to a boundary pixel.
Then, the inverse of the distance map $\Phi$ is used as the weight for the cross-entropy loss so that pixels close to the boundary get a higher weight and pixels away from the boundary get a lower weight.

\begin{equation}
L_{DMCE}(y, t, \Phi) = -\sum_{n=1}^{N} (1+\Phi_n) \log (t_n \cdot y_n)
\label{eqn: Distance map}
\end{equation}
The constant 1 is added to $\phi$ to avoid the vanishing gradient problem.

\subsection{Region-level}
Region-level loss functions take a broader view in semantic segmentation tasks. Instead of focusing on each pixel, these methods prioritize the overall accuracy of object segmentation. Their goal is to ensure that the predicted segmentation mask closely matches the higher-level ground truth mask, capturing the essence of the object's shape and layout. Region-level loss functions are particularly valuable when the global context and object completeness are more important than pixel-level accuracy.

\subsubsection{Dice Loss}
The Dice loss originates from the Dice Coefficient which is a measure of similarity between two sets of data.
It is commonly used in image segmentation to evaluate the overlap between a predicted segmentation mask and the target segmentation mask.
It is defined as the size of the intersection of the predicted segmentation mask and the ground truth segmentation mask, divided by their sum.
It is calculated separately for each class and the average is reported.
It can be written as:
\begin{equation}
\text{Dice Coefficient}=\frac{2 |Y \cap T|}{|Y| + |T|}
\end{equation}
where Y is the binary segmentation prediction mask and T is the binary segmentation target mask for a single class.
The dice coefficient is commonly used in semantic segmentation because it is easy to compute, provides a single-value summary of performance, and achieves a good balance between precision and recall.
It is especially useful when the object of interest is small or rare and the class distribution is imbalanced.

The Dice loss was proposed by Milletari et al.~\cite{milletari2016fully} as Eqn. \ref{eqn:Dice}.
It can be viewed as a relaxed, differentiable Dice Coefficient.
\begin{equation}
L_{dice} = 1 - \frac{1}{C} \sum_{c=0}^{C-1} \frac{2 \sum_{n=1}^{N}{t_n^c y_n^c}} {\sum_{n=1}^{N}{(t_n^c + y_n^c)}}
\label{eqn:Dice}
\end{equation}
It is computed separately for each target class and the average over all classes is used.
The predictions are not taken as certain {0,1} but relaxed and taken as probabilities [0,1].
This makes the loss function become differentiable and can be optimized using gradient descent approaches.
Finally, the relaxed Dice Coefficient is subtracted from 1 to make it a loss function to be minimized rather than maximized.
This is a popular choice for imbalanced datasets because it prevents the model from ignoring the minority classes by focusing on the overlapping regions between the predicted and ground truth masks.

\subsubsection{Log-Cosh Dice Loss}
Jadon~\cite{Jadon2020ASO} wrapped the Dice Loss with a log-cosh function, which is defined as:
\begin{equation}
    L_{lc-dice} = \log (\text{cosh} (DiceLoss))
\end{equation}
with $\text{cosh}(x) = \left( \text{e}^x - \text{e}^{-x}\right)/2$.
The derivative of the log-cosh function, tanh, is a smooth function ranging within $\pm 1$.

The Log-Cosh Dice Loss offers several key advantages in segmentation tasks. Firstly, it enhances smoothness and robustness to outliers, mitigating the influence of noisy annotations or image artefacts. This feature ensures more stable training, particularly in situations prone to data irregularities. Secondly, the loss's inherent smoothness fosters improved optimization, preventing the destabilizing impact of sharp gradients often encountered in conventional Dice Loss. This proves especially beneficial when employing gradient-based optimization methods like SGD. Lastly, the Log-Cosh Dice Loss strikes a balance between precision and recall, addressing a limitation of the Dice Loss, which typically emphasizes precision over recall. This equilibrium stems from its smoothness, potentially yielding superior segmentation results. In a binary segmentation example, while the Dice Loss penalizes false positives heavily, the Log-Cosh Dice Loss provides a more even-handed approach, smoothing the loss landscape, reducing sensitivity to outliers, and ultimately facilitating better management of class imbalance and a focus on enhancing both precision and recall.

\subsubsection{Generalised Wasserstein Dice Loss}
The Wasserstein distance, also known as the Earth Mover's Distance (EMD) is the distance between two probability distributions calculated by using the minimum cost of transforming one distribution into the other.
From this definition, Wasserstein distance requires finding an "optimal transport" that minimizes the cost of moving from one distribution to the other.
When the number of possible paths is finite, this minimization problem can be formulated as a linear programming problem.

In the context of semantic segmentation, Fidon et. al.~\cite{fidon2018generalised} proposed to use Wasserstein distance to calculate a loss term depending on predicted $y_n$ and target $t_n$ class probabilities.
In this approach, the transition cost between various classes is depicted through a matrix denoted as $M_{CxC}$, allowing to impose a milder penalty on errors between semantically similar classes, such as "left-kidney" and "right-kidney" and vice versa.
As a result, the loss function can be designed to take the inter-class relationships into account.

\begin{equation}
\text{Generalized Wasserstein Dice Loss} = -\sum_{n=1}^{N} \sum_{c=1}^{C} M_{c,c} \cdot t_n^c \cdot y_n^c
\end{equation}

where, $y_n$ represents the predicted class probabilities for the $n^\text{th}$ pixel.
$t_n$ represents the one-hot encoding vector representing the target class of the $n^\text{th}$ pixel.
$M_{CxC}$ is a matrix representing transition costs between various classes, allowing for the consideration of semantic similarity between classes.

\subsubsection{IOU (Jaccard) Loss}
IOU loss originates from the Intersection over Union (IoU) metric, which is also known as the Jaccard index.
It is defined as the size of the intersection of the predicted segmentation mask and the ground truth segmentation mask, divided by the size of their union.
\begin{equation}
\text{Intersection over Union}=\frac{|Y \cap T|}{|Y \cup T|}
\end{equation}
Similar to the Dice coefficient, it is calculated for each class and the mean (mIoU) is used.

IoU loss was proposed by Rahman et. al.~\cite{Rahman2016OptimizingII}.
It can be viewed as a relaxed, differentiable mIoU.
\begin{equation}
L_{IoU} = 1 - \frac{1}{C} \sum_{c=0}^{C-1} \frac{\sum_{n=1}^{N}{t_n^c y_n^c}} {\sum_{n=1}^{N}{(t_n^c + y_n^c - t_n^c y_n^c)}}
\label{eqn:IoU}
\end{equation}

\subsubsection{Lovász-Softmax Loss}
The Lovász-Softmax loss~\cite{Berman2017TheLL} is a function used as a surrogate for directly optimizing the IoU metric.
The basic idea behind this is to treat the predicted scores of each class as a set of ordered values, and then define a function that measures the discrepancy between these ordered values and the ground truth ordering.
This discrepancy is then used as the loss to be minimized during training.
It has been shown to have better mIoU scores compared to training with cross-entropy loss.

\begin{equation}
L_{Lovasz} = \frac{1}{C} \sum_{c=0}^{C-1} \overline{\Delta_{J_c}}(\boldsymbol{m}(c))
\label{eqn:Lovász}
\end{equation}

where $\overline{\Delta_{J_c}}(\boldsymbol{m}(c))$ is Lovász hinge applied to the IoU (Jaccard) loss calculated using the hinge losses.

\subsubsection{Tversky Loss}
Tversky Loss originates from the Tversky index~\cite{Tversky1977FeaturesOS} which is an asymmetric similarity measure between two sets of data.
It is a generalization of the Dice coefficient and IoU that allows weighing false positives and false negatives independently.
It is defined as:
\begin{equation}
\text{Tversky index}=\frac{|Y \cap T|}{|Y \cap T|+\alpha|Y \backslash T|+\beta|Y \backslash T|}
\end{equation}
where $\alpha$ and $\beta$ are the weights for false negatives and false positives.
When $\alpha = \beta = 0.5$ it reduces to the Dice coefficient and when $\alpha = \beta = 1$ it reduces to the IoU.

Inspired by the Tversky index, Tversky loss~\cite{salehi2017tversky} is proposed:
\begin{equation}
L_{T} = 1 - \frac{1}{C} \sum_{c=0}^{C-1} \frac{\sum_{n=1}^{N}{t_n^c y_n^c}} {\sum_{n=1}^{N}{t_n^c y_n^c + \alpha t_n^c (1-y_n^c) + \beta y_n^c (1-t_n^c))}}
\label{eqn:Tversky}
\end{equation}

\subsubsection{Focal Tversky Loss} \label{sec:FocalTversky}
Similar to Focal loss, Focal Tversky loss~\cite{Abraham2018ANF} increases the weight of hard-to-classify pixels.
\begin{equation}
L_{FT}=\sum_{c=0}^{C-1}\left(L_T^c\right)^{1 / \gamma}
\end{equation}
where $L_T^c$ represents the Tversky loss of class c.
Focal Tversky loss is identical to Tversky loss for $\gamma = 1$.
Abraham et. al. recommended a range of [1,3] for $\gamma$ that makes the model focus on misclassified pixels.
However, when the training is close to convergence Focal Tversky loss becomes suppressed and prevents the model from pushing for complete convergence.

\subsubsection{Sensitivity Specificity Loss}
The Sensitivity and Specificity terms are widely used in evaluating the performance of machine learning models.
Sensitivity, also known as recall, is the ratio of correctly classified positive predictions to actually positive samples.
Specificity is the proportion of true negative samples that were classified as negative.
The two terms are defined as follows : 
\begin{equation}
\text{Sensitivity}=\frac{|Y \cap T|}{|T|}
\end{equation}
\begin{equation}
\text{Specificity}=\frac{|Y' \cap T'|}{|T'|}
\end{equation}

To control the trade-off between FNs and FPs in case of imbalanced data, the Sensitivity-specificity loss~\cite{Brosch2015DeepCE} was devised.
Defined in equation \eqref{SSL}, it struggles to adjust the weights assigned to FNs and FPs using the w parameter in equation \eqref{SSL}

\begin{equation}
L_{SSL}= 1 - \alpha * \text{sensitivity}+(1-\alpha) * \text{specificity}
\label{SSL}
\end{equation}

\begin{equation}
\text{sensitivity} =\frac{1}{C} \sum_{c=0}^{C-1} \frac{\sum_{n=1}^{N}{t_n^c (t_n^c - y_n^c)^2 }} {\sum_{n=1}^{N}{t_n^c}}
\label{eqn:sensitivity}
\end{equation}

\begin{equation}
\text{specifity} =\frac{1}{C} \sum_{c=0}^{C-1} \frac{\sum_{n=1}^{N}{(1-t_n^c) (t_n^c - y_n^c)^2 }} {\sum_{n=1}^{N}{(1-t_n^c)}}
\label{eqn:specifity}
\end{equation}

\subsubsection{Region Mutual Loss (RMI)}
Although cross-entropy based loss functions are effective for pixel-wise classification, they overlook interdependencies between pixels within an image.
This limitation has encouraged the exploration of alternative methods, including approaches based on conditional random fields~\cite{Lafferty2001ConditionalRF} and pixel affinity.
While these techniques hold promise for capturing pixel relationships, they often require longer computation times, are sensitive to variations in visual attributes, and require additional memory resources.
Region Mutual Information (RMI) loss~\cite{Zhao2019RegionMI} is proposed to address the inherent limitations of conventional pixel-wise loss functions by exploiting the interdependencies that exist between pixels in an image.
RMI is based on Mutual Information (MI), which is defined between two random variables to quantify the information obtained about one variable by observing the other.
It considers each pixel with its 8 neighbours to represent it, so that each pixel of an image is a 9-dimensional (9-D) point. In other words, each image is transformed into a multi-dimensional distribution of these 9-D points. Finally, the similarity between the multi-dimensional distributions of the ground truth and the prediction of the model is maximized with the help of mutual information (MI). 
Instead of calculating MI between these multi-dimensional distributions, they proposed to calculate a lower bond of MI between them. Also, before constructing these multi-dimensional distributions, they use a down-sampling strategy to reduce the additional memory consumption.

The simplified lower bound of MI is written as Eqn. \ref{eqn:RMI}:

\begin{equation}
I_l(Y;P) = - \frac{1}{2}log(det(\Sigma_{Y|P})),
\label{eqn:RMI}
\end{equation}

\noindent where $\Sigma_{Y|P}$ is the posterior covariance of $Y$ given $P$. For more details about how to approximate the posterior covariance of $Y$, refer to the main article~\cite{Zhao2019RegionMI}.

\subsubsection{Robust T-Loss}
The Robust T-Loss~\cite{gonzalez2023robust} takes a unique approach to segmentation by emphasizing robustness.
It does this by using the negative log-likelihood from the Student-t distribution, which is known to handle noisy data and outliers very well.
This distribution is characterized by its tails being "heavier" than the more common normal distribution.
These heavy tails make the Student-t distribution great at handling data points that are far from the usual pattern.

In regular loss functions, we often use the Mean Squared Error (MSE), which comes from the negative log-likelihood of the normal distribution.
The RTL changes things by replacing the normal distribution with the Student-t distribution.
\begin{equation}
L_{RT} =\frac{1}{N} \sum_{i=1}^N-\log p\left(\mathbf{y}_i \mid \boldsymbol{\Sigma} ; \nu\right)
\end{equation}
Here, $p\left(\mathbf{y}_i \mid \boldsymbol{\Sigma} ; \nu\right)$ is the probability based on Student-t distribution
This change makes the loss function much more resistant to the influence of noisy labels and outliers.

The Robust T-Loss has a key parameter $\nu$, which controls how the loss function responds to different levels of noise.
When $\nu$ is low, the loss is similar to the MSE, and at high values, it is similar to the Mean Absolute Error (MAE).
A significant advantage of the Robust T-Loss is its ability to learn the optimal tolerance level for label noise during the training process.
This distinguishes it from other methods that require prior knowledge of the noise level or complex computations.
By directly integrating the adaptation process into backpropagation, the loss function essentially teaches itself how to handle noisy labels, eliminating the need for additional computations.

\subsection{Boundary-level}
Boundary-Level Loss Functions: Boundary-level loss functions specialize in the precision of object boundaries within the segmentation task. Their primary objective is to sharpen object boundaries and effectively separate overlapping objects. These loss functions work by minimizing the distance or dissimilarity between the predicted object boundaries and the ground truth boundaries. They are useful in tasks where distinguishing object boundaries is critical, such as image inpainting or scene segmentation.

\subsubsection{Boundary Loss}
The Boundary Loss, as introduced by Kervadec et al.~\cite{Kervadec2018BoundaryLF} in their work, presents an innovative approach for addressing imbalanced segmentation tasks, particularly when the size of the foreground region significantly contrasts with that of the background region. This imbalance frequently results in performance deterioration and training instability when conventional regional loss functions, such as the Dice loss, are applied. The Boundary Loss adeptly confronts these challenges by centring its focus on the boundary regions.

The crux of the Boundary Loss lies in its utilization of a distance metric tailored to boundaries. This metric serves to quantify the disparities between predicted boundaries and their corresponding ground truth representations, encapsulating variations normal to the ground truth boundary direction. The L2 distance plays a foundational role in assessing changes in boundaries, and it is mathematically defined as follows:

\begin{equation}
L_{B} = \int_{\Omega} \phi_G(q) s(q) dq
\end{equation}

Here, in the equation above, $s(q)$ denotes the probability predictions generated by the model, and $\phi_G(q)$ represents the distance term. However, it is essential to acknowledge that directly incorporating this distance metric as a loss function is a formidable task. This complexity primarily stems from the challenge of representing boundary points as differentiable functions, which are derived from the outputs of neural networks. Consequently, researchers often shy away from employing boundary-based loss functions due to this intricate issue. To surmount this limitation, the authors of Boundary Loss draw inspiration from insights in discrete optimization techniques, which are traditionally employed in the context of curve evolution.

\subsubsection{Hausdorff Distance Loss}
Hausdorff Distance (HD) is a common evaluation metric used in medical image segmentation.
The Hausdorff Distance is a metric defined on pairs of sets and it quantifies the maximum distance between points in one set to the nearest point in the other set, capturing the worst-case scenario.
In this context, two non-empty point sets are considered, denoted as \(X\) and \(Y\), with a distance measure, denoted as \(d(x, y)\), between points \(x \in X\) and \(y \in Y\), often using metrics such as Euclidean or Manhattan distances.
The Hausdorff distance is defined as:

\begin{equation}
\operatorname{HD}(X, Y)=\max (\operatorname{d_h}(X, Y), \operatorname{d_h}(Y, X))
\label{Hausdorff}
\end{equation}

\begin{equation}
\operatorname{d_h}(X, Y)=\max _{x \in X} \min _{y \in Y}d(x,y),
\end{equation}

\begin{equation}
\operatorname{d_h}(Y, X)=\max _{y \in Y} \min _{x \in X}d(y,x).
\end{equation}

In the case of image segmentation, HD is calculated between the boundaries of the predicted and ground truth masks.
Despite being a common metric, HD has its drawbacks.
Unlike other metrics that use the overall segmentation performance, HD solely depends on the largest error and it is overly sensitive to outliers.
As a result, optimizing solely to minimize the largest error can lead to algorithm instability and unreliable results.
Moreover, minimizing only the largest segmentation error can degrade the overall segmentation performance, especially in the case of compound shapes, which are common in medical imaging.
This is because, although the model may be able to achieve sufficient accuracy over most parts of the image, it may encounter large errors in a few exceptionally difficult regions.
Karimi et al.~\cite{Karimi2019ReducingTH} proposed an approach to directly optimize neural networks for the reduction of HD.
They propose three different loss functions to minimize HD by taking three different approaches to approximate it in a differentiable manner.
They show the potential of these loss functions to reduce high errors while not degrading the overall segmentation performance.

\subsubsection{Boundary-aware Loss}
Hayder et al.~\cite{hayder2017boundary} proposed a boundary-aware loss in the domain of instance-level semantic segmentation.
The idea is to predict a pixel-wise distance map instead of a binary foreground mask.
This distance map represents either the distance to the nearest object boundary (if inside the object) or its background state.
To ensure consistency across different object shapes and sizes, the distance values are first normalized and truncated within a specified range.
\begin{equation}
    D(p) = \min \left( \min_{\forall q \in Q} \lceil d(p,q) \rceil, R \right),
\end{equation}
\noindent where $d(p,q)$ calculates the Euclidean distance between a pixel $p$ and a boundary pixel $q$.
The maximum distance $D(p)$ is ceiled with $\lceil \cdot \rceil$ and thresholded with $R$ to generate a truncated map.
Then, quantizing these distance values into uniform histogram bins converts the distance map into a set of binary maps.
This converts the problem into $K$ binary segmentation tasks, which are solved with $K$ binary cross-entropy losses.
During inference, the pixel-wise predicted distances are used to create disks with their corresponding pixel in the centre.
The union of these disks results in segmentation masks.

\subsubsection{Active Boundary Loss}
Active boundary loss~\cite{wang2022active} is designed to specifically supervise and enhance predicted boundaries during training.
Here, boundary information is embedded into the training process, making it possible for the network to pay special attention to boundary pixels.
First, predicted boundary pixels are identified by calculating a boundary map using the KL divergence of neighbouring pixels.
This map highlights pixels that are likely to be part of object boundaries.
Then, for each predicted pixel, a target direction towards the nearest ground truth boundary is calculated.
This direction is encoded as a one-hot vector, allowing the representation of pixel movement in a probabilistic manner.
The cross-entropy loss is then calculated based on the predicted directions, encouraging the network to align the predicted and the ground-truth boundaries.
\begin{equation}
L_{AB}=\frac{1}{N_b} \sum_i^{N_b} {\Lambda}\left(\mathbf{M}_i\right) \mathbf{C E}\left(\mathbf{D}_i^p, \mathbf{D}_i^g\right)
\end{equation}
Here, ${\Lambda}\left(\mathbf{M}_i\right)$ is the weight function and $\mathbf{C E}\left(\mathbf{D}_i^p, \mathbf{D}_i^g\right)$ is the cross-entropy over the neighboring pixels.
This dynamic behavior ensures that the predicted boundaries continuously adapt and align themselves with the evolving ground-truth boundaries as the network parameters are updated during training.

\subsubsection{InverseForm Loss}
Borse et al.~\cite{Borse2021InverseFormAL} developed the InverseForm loss which focuses on boundary transformation between the predicted and ground truth objects.
This helps to assign a lower loss to predictions that do not perfectly align with the ground truth but have structural similarities to it.
First, they train an MLP called inverse transformation network that takes two boundary maps as input and predicts a transformation matrix $\hat{\theta}$ between them.
For instance, for two boundary maps that match perfectly, the network should output an identity matrix as their relative transformation.
After training this inverse transformation network, they freeze its weights and use it to calculate the loss for a segmentation model.
Specifically, they compute the Euclidean or Geodesic distances between the identity matrix and predicted transformation matrix $\hat{\theta}$ and combine it with the cross-entropy loss as shown below:

\begin{equation}
L_{total}= L_{CE}\left(y_{pred}, y_{gt}\right)+\beta \\ L_{BCE}\left(b_{pred}, {b_{gt}}\right) \\
+\gamma L_{IF}\left(b_{pred}, b_{gt}\right)
\end{equation}

Here, $L_{CE}$ and $L_{BCE}$ calculate the cross-entropy loss for the whole mask and for its boundary pixels while $L_{if}$ represents the InverseForm loss.
$y_pred$ and $y_gt$ denote predicted and ground truth segmentation masks, $b_pred$ and $b_gt$ denotes the corresponding boundaries.
$L_{BCE}$ and $L_{IF}$ are scaled by constants $\beta$ and $\gamma$ for controlling the impact of their respective losses.

\subsubsection{Conditional Boundary Loss}

In order to enhance boundary performance, Wu et al.~\cite{Wu2023ConditionalBL} suggest a conditional boundary loss (CBL) that establishes a distinct optimization objective for each boundary pixel, dependent on its neighbouring context and enhances intra-class consistency, inter-class separation, and boundary precision by aligning each pixel with its class center and filtering out noise. This is achieved through a simple yet effective sampling strategy named the conditional correctness-aware sampling (CCAS) strategy, which only selects the correctly classified same-class neighbours as the positive samples of a boundary pixel and the correctly classified different-class neighbours as the negative samples.
The proposed CBL consists of two terms: the $A2C$ (pairs between
an anchor and its uniquely generated local class centre) loss term and the $A2P\&N$ (pairs between an anchor and its selected positive and negative samples) loss term. The $A2C$ loss term supervises the distance between each boundary pixel and its corresponding local class centre that is generated from the correctly classified surrounding neighbours. The $A2P\&N$ loss term supervises the similarity between the boundary pixel and its positive and negative samples, which are selected using the CCAS strategy. The CBL is then combined with the commonly used cross-entropy (CE) loss to form the overall training loss, which is used to optimize the segmentation network during end-to-end training.

\subsubsection{Boundary Difference Over Union Loss}
Sun et al.~\cite{sun2023boundary} proposed the boundary difference over union (Boundary DoU) loss which aims to improve the segmentation quality at object boundaries.
Here, the boundary of the objects is defined as $d$ outermost pixels.
It is inspired by the boundary IoU metric, which is the IoU when only the boundary regions are considered for both prediction and target.
The loss is formulated as follows:
\begin{equation}
L_{BDoU}=\frac{G \cup P - G \cap P}{G \cup P - \alpha * G \cap P}
\end{equation}
Here, $\alpha$ is a weighting term controlling the importance of the boundary region.
For a relatively larger object, the boundary pixels constitute a small portion of the whole area resulting in a low loss even if only the inner part is segmented correctly.
For such a case, $\alpha$ should be closer to 1, indicating that boundary pixels are assigned a higher significance in relation to the internal region.
On the contrary, for smaller objects $\alpha$ should be closer to 0, converging to the IoU loss.
To ensure these authors propose the weighting term $\alpha=1-\frac{2C}{S}, \alpha \in[0,1)$ where C denotes the circumference and S denotes the area of the object.
As a result, they ensure precise segmentation even for the boundaries of large objects in the image.

\subsubsection{Region-wise Loss}
The core concept of region-wise (RW) loss~\cite{VALVERDE2023109208} is combining the softmax probability values with RW maps.
The RW map is defined for each pixel in an image and for each class label.
It influences how much a particular pixel's prediction should contribute to the loss calculation based on its class label and its location within the image.
\begin{equation}
L_{RW}=\sum_{i=1}^N \hat{\boldsymbol{y}}_i^{\top} \boldsymbol{z}_i
\end{equation}
where $\hat{\boldsymbol{y}}_i^{\top}$ is the softmax of the predictions and $\boldsymbol{z}_i$ is the RW map values at that pixel.
Different types of RW maps can be designed based on the specific requirements of the segmentation task.
For example, RW-Boundary maps use the Euclidean distance transform to create distance-based maps that highlight boundaries between classes.
This framework provides a versatile and unified approach to tackle class imbalance and pixel importance simultaneously
Moreover, the paper demonstrates the adaptability of the RW loss framework by reformulating well-known loss functions like Boundary loss and Active Contour loss.
This not only provides new insights into the relationships between these loss functions but also establishes the versatility of the RW loss framework.
They delve further into the optimization stability of RW maps and introduce the concept of rectified region-wise (RRW) maps.
These RRW maps address optimization instability concerns, thereby enhancing the convergence and stability of the training process.
Through empirical evaluations across various segmentation tasks, the paper showcases the efficacy of RRW maps.

\subsection{Combination}
The combo approach harmonizes elements from the three distinct categories (pixel-level, region-level, and boundary-level) to optimize semantic segmentation performance. By integrating multiple loss functions, this approach seeks equilibrium between pixel-wise precision, overall object segmentation quality, and boundary delineation accuracy. The combo approach offers versatility and adaptability, leveraging the strengths of each category to address the specific challenges presented by diverse segmentation tasks and dataset characteristics.

\subsubsection{Combo Loss}
In semantic segmentation, the most common practice is to combine Dice loss and weighted cross-entropy loss within the Combo loss~\cite{Taghanaki2018ComboLH} to overcome the class imbalance problem.
Here, weighted cross-entropy loss allows for overcoming the data imbalance problem by giving more weight to underrepresented classes while Dice loss allows for the segmentation of smaller objects.
Furthermore, weighted cross-entropy loss provides smooth gradients while Dice loss helps avoid local minima.
It simply adds the cross-entropy loss and the Dice loss using a modulating term to control the contribution of each loss function, the overall equation is defined as :

\begin{equation}
L_{combo} =\alpha L_{WCE} + (1-\alpha) L_{dice}
\label{eqn:combo}
\end{equation}

where $\alpha$ controls the weight of Dice loss compared to weighted cross-entropy loss, and weights of the cross-entropy control the amount of model punishment for different target classes.

\subsubsection{Exponential Logarithmic Loss}
Exponential Logarithmic Loss~\cite{Wong20183DSW}, similar to the Combo loss, combines weighted cross-entropy loss and Dice loss to overcome the class imbalance problem.
The difference is exponential logarithmic loss takes the logarithm and exponential of both of the loss functions before combining them.
This gives the flexibility to control how much the model focuses on easy/hard pixels.
The proposed loss function is defined as follows:

\begin{equation}
L_{\text{Exp-Log}} = \alpha L_{\text{Exp-Log-Dice}} + \beta L_{\text{Exp-Log-WCE}}
\end{equation}

where $L_{\text{Exp-Log-Dice}}$ is the exponential logarithmic Dice loss and $L_{\text{Exp-Log-WCE}}$ is the exponential logarithmic weighted cross-entropy loss:

\begin{equation}
L_{\text{Exp-Log-Dice}} = (-\log (L_{Dice}))^{\gamma_{\text{Dice}}}
\end{equation}

\begin{equation}
L_{\text{Exp-Log-WCE}} = (-\log (L_{WCE}))^{\gamma_{\text{WCE}}}
\end{equation}

Here $\gamma_{\text{WCE}}$ and $\gamma_{\text{WCE}}$ can be used to control the focus of the loss function.
Specifically, with $\gamma > 1$, the loss focuses more on hard-to-classify pixels and vice versa.

\subsubsection{Unified Focal Loss}
The Unified Focal loss \cite{yeung2022unified} is another loss function designed to address the class imbalance by combining the Focal loss and Focal Tversky loss.
It mitigates the issues associated with loss suppression and over-enhancement during training.
This is achieved through unifying similar hyperparameters.
\begin{equation}
L_{UF}=\lambda L_{F}+(1-\lambda) L_{FT}
\end{equation}

The Unified Focal loss generalizes common loss functions like Dice and cross-entropy, making them special cases within its framework.
Importantly, by reducing the hyperparameter search space it strikes a balance between simplicity and efficacy, simplifying optimization while maintaining its effectiveness.
Experimental results support its advantages, making it a powerful tool for training models robust to class imbalance.

\begin{figure*}[ht]
    \centering
    \includegraphics[width=\textwidth]{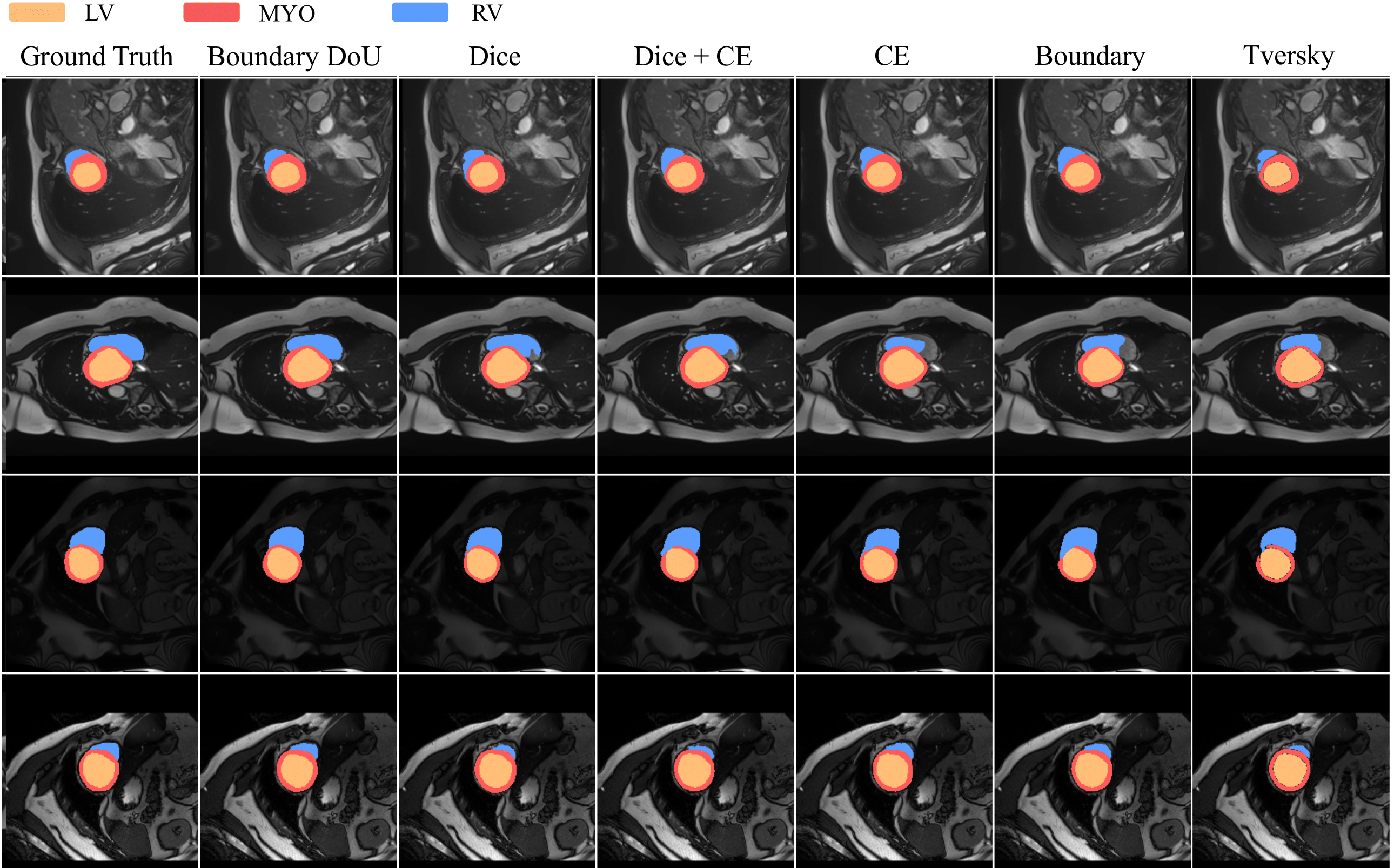}
    \caption{The qualitative comparison of segmentation results on the ACDC dataset from~\cite{sun2023boundary}. (Row 1\&2: TransUNet, and Row 3\&4: UNet)} \label{acdc}
\end{figure*}

\section{Discussion}
Table \ref{tab:highlights} summarizes the advantages, disadvantages and use cases of the discussed loss functions ordered based on their category, i.e., Pixel-Level, Region-Level, Boundary-Level or Combination. Pixel-level losses have the advantage of handling imbalanced class distributions by either considering every pixel, shifting the focus to difficult-to-segment pixels, or penalizing segmentation errors. Since the focus is on global statistics this can cause softer segmentation boundaries compared to other losses. Region-level losses compute overlaps or similarities between segmentation regions, often associated with semantic segmentation performance measures, to guide the network toward better performance. More advanced losses can exploit the tradeoff between false positives and false negatives and are more robust to outliers and noisy labels. Some region-based losses encounter problems during optimization, either with unstable gradients or with losses that are not fully differentiable. Losses based on boundary considerations generally focus on a sharp segmentation boundary to obtain a better segmentation mask. These losses have various limitations, i.e., they are limited to binary segmentation problems, exploding gradients during optimization, or problems with compound shapes. Combo Losses try to combine the advantages of different losses or mitigate their limitations. Therefore, their advantages and disadvantages mostly depend on the underlying losses. In general, all losses, including hyperparameters, are sensitive to this choice, as it can have a large impact on performance, which is proven in our experiments. Usually, there are no general guidelines for choosing hyperparameter settings, since the optimal choice depends on the data and the task. This requires extensive experimentation, especially for losses with multiple hyperparameters, to find the optimal setting for maximum performance.

\begin{table*}
\centering
\caption{A summary table of the presented loss function in order of the proposed taxonomy. Advantages, disadvantages and use cases are presented.}
\label{tab:highlights}
\fontsize{11pt}{21pt}
\resizebox{\textwidth}{!}{%
\begin{tabular}{|c|llll|}

\rowcolor[rgb]{0.753,0.753,0.753}
\textbf{Category} 
& \multicolumn{1}{c|}{\textbf{Loss Function}}
& \multicolumn{1}{c|}{\textbf{Advantages}}
& \multicolumn{1}{c|}{\textbf{Disadvantages}} 
& \multicolumn{1}{c|}{\textbf{Use Cases}}
\\ 



\rowcolor[rgb]{0.882,0.98,0.882} {\cellcolor[rgb]{0.722,0.957,0.722}}
& Cross-Entropy Loss
& \begin{tabular}[c]{@{}>{\cellcolor[rgb]{0.882,0.98,0.882}}l@{}}
$\bullet$ Variations allow handling imbalanced datasets
\\ $\bullet$ Logarithm operation strengthens numerical stability
\\$\bullet$ Only predicted probability of target class effects loss\end{tabular}    
& $\bullet$ Focus on global statistics can cause softer segmentation 
& $\bullet$ General semantic segmentation tasks
\\
\rowcolor[rgb]{0.882,0.98,0.882} {\cellcolor[rgb]{0.722,0.957,0.722}} & & & &
\\
\rowcolor[rgb]{0.882,0.98,0.882} {\cellcolor[rgb]{0.722,0.957,0.722}}
& TopK Loss
& \begin{tabular}[c]{@{}>{\cellcolor[rgb]{0.882,0.98,0.882}}l@{}}
$\bullet$ Focus on the most difficult pixels improves performance in initial weak regions
\\$\bullet$ Handling of class imbalance by focusing on most challenging samples\end{tabular}
& \begin{tabular}[c]{@{}>{\cellcolor[rgb]{0.882,0.98,0.882}}l@{}}
$\bullet$ Selection of 'K' parameter impacts performance leading to \\ hyperparameter sensitivity
\\$\bullet$ Additional computational cost to identify top-K uncertain predictions\end{tabular}   
& \begin{tabular}[c]{@{}>{\cellcolor[rgb]{0.882,0.98,0.882}}l@{}}
$\bullet$ General semantic segmentation tasks
\\$\bullet$ Overlapping objects\end{tabular}
\\
\rowcolor[rgb]{0.882,0.98,0.882} {\cellcolor[rgb]{0.722,0.957,0.722}} & & & &
\\
\rowcolor[rgb]{0.882,0.98,0.882} {\cellcolor[rgb]{0.722,0.957,0.722}}
& Focal Loss
& \begin{tabular}[c]{@{}>{\cellcolor[rgb]{0.882,0.98,0.882}}l@{}}
$\bullet$ Handling of highly-imbalanced dataset
\\$\bullet$ Improves difficult and misclassified examples\end{tabular}
& $\bullet$ Additional hyperparameter of the focal factor that needs to be tuned
& \begin{tabular}[c]{@{}>{\cellcolor[rgb]{0.882,0.98,0.882}}l@{}}
$\bullet$ Highly-imbalanced datasets
\\ $\bullet$ Complex scenes with challenging objects\end{tabular}
\\
\rowcolor[rgb]{0.882,0.98,0.882} {\cellcolor[rgb]{0.722,0.957,0.722}} & & & &
\\
\rowcolor[rgb]{0.882,0.98,0.882} \multirow{-11}{*}{{\cellcolor[rgb]{0.722,0.957,0.722}}
\textbf{Pixel Level}}
& Distance map derived cross-entropy loss
& \begin{tabular}[c]{@{}>{\cellcolor[rgb]{0.882,0.98,0.882}}l@{}}
$\bullet$ Penalized segmentation errors lead to sharper segmentation boundaries
\\$\bullet$ Boundary awareness during training \end{tabular}
& \begin{tabular}[c]{@{}>{\cellcolor[rgb]{0.882,0.98,0.882}}l@{}}
$\bullet$ Computation of distance maps adds overhead to the training process    
\\ $\bullet$ Dependent on the hyperparameter decay rate for distance weighting \end{tabular} 
& \begin{tabular}[c]{@{}>{\cellcolor[rgb]{0.882,0.98,0.882}}l@{}}
$\bullet$ Hard to segment boundaries
\\$\bullet$ Medical image segmentation\end{tabular}
\\ 
\hline


\rowcolor[rgb]{0.929,0.867,0.929} {\cellcolor[rgb]{0.835,0.675,0.835}}
& Dice Loss
& \begin{tabular}[c]{@{}>{\cellcolor[rgb]{0.929,0.867,0.929}}l@{}}
$\bullet$ Originates from the dice coefficient which is a common performance metric
\\$\bullet$ Model considers minority classes with a focus on overlapping regions
\\$\bullet$ Good balance between precision and recall\end{tabular}
& \begin{tabular}[c]{@{}>{\cellcolor[rgb]
{0.929,0.867,0.929}}l@{}}
$\bullet$ Unstable gradients especially for minor classes with a small \\gradient denominator
\\$\bullet$ Relaxed formulation to make the loss differentiable and \\suitable for optimization
\\$\bullet$ No consideration of inter-class relationships\end{tabular}
& \begin{tabular}[c]{@{}>{\cellcolor[rgb]{0.929,0.867,0.929}}l@{}}
$\bullet$ Medical image segmentation
\\$\bullet$ Small or rare objects
\\$\bullet$ Imbalanced class distribution\end{tabular}
\\
\rowcolor[rgb]{0.929,0.867,0.929} {\cellcolor[rgb]{0.835,0.675,0.835}} & & & &
\\
\rowcolor[rgb]{0.929,0.867,0.929} {\cellcolor[rgb]{0.835,0.675,0.835}}
& Log-Cosh Dice Loss
& \begin{tabular}[c]{@{}>{\cellcolor[rgb]{0.929,0.867,0.929}}l@{}}
$\bullet$ Smooth function while gradients remain continuous and finite
\\$\bullet$ Inherits advantages from Dice Loss\end{tabular}
& $\bullet$ No consideration of inter-class relationships     
& \begin{tabular}[c]{@{}>{\cellcolor[rgb]{0.929,0.867,0.929}}l@{}}
$\bullet$ Medical image segmentation
\\$\bullet$ Small or rare objects
\\$\bullet$ Imbalanced class distribution\end{tabular}
\\
\rowcolor[rgb]{0.929,0.867,0.929} {\cellcolor[rgb]{0.835,0.675,0.835}} & & & &
\\
\rowcolor[rgb]{0.929,0.867,0.929} {\cellcolor[rgb]{0.835,0.675,0.835}}
& Generalised Wasserstein Dice Loss
& \begin{tabular}[c]{@{}>{\cellcolor[rgb]{0.929,0.867,0.929}}l@{}}
$\bullet$ Wasserstein distance measures the minimum cost to transform one distribution into another \\ resulting in a linear programming problem
\\$\bullet$ Inter-class relationships improve segmentation performance\end{tabular}    
& $\bullet$ Computational overhead during training  
& $\bullet$ Data with semantically similar classes
\\
\rowcolor[rgb]{0.929,0.867,0.929} {\cellcolor[rgb]{0.835,0.675,0.835}} & & & &
\\
\rowcolor[rgb]{0.929,0.867,0.929} {\cellcolor[rgb]{0.835,0.675,0.835}}
& IOU (Jaccard) Loss
& \begin{tabular}[c]{@{}>{\cellcolor[rgb]{0.929,0.867,0.929}}l@{}}
$\bullet$ Overlap-based loss is directly connected to segmentation performance\\
$\bullet$ Class imbalance handled by agreement between prediction and ground truth\\
$\bullet$ Sharp segmentation boundaries due to overlap loss\end{tabular}
& $\bullet$ Vanishing gradient problem in region of minimal overlap 
& \begin{tabular}[c]{@{}>{\cellcolor[rgb]{0.929,0.867,0.929}}l@{}}
$\bullet$ General semantic segmentation
\\$\bullet$ Medical image segmentation
\\$\bullet$ Imbalanced class distributions\end{tabular}
\\
\rowcolor[rgb]{0.929,0.867,0.929} {\cellcolor[rgb]{0.835,0.675,0.835}} & & & &
\\
\rowcolor[rgb]{0.929,0.867,0.929} {\cellcolor[rgb]{0.835,0.675,0.835}}
& Lovász-Softmax loss
& \begin{tabular}[c]{@{}>{\cellcolor[rgb]{0.929,0.867,0.929}}l@{}}
$\bullet$ Related to IoU performance measure leading to sharper segmentation boundaries\\
$\bullet$ Handles class imbalance by penalizing the measured discrepancy \end{tabular}
& \begin{tabular}[c]{@{}>{\cellcolor[rgb]{0.929,0.867,0.929}}l@{}}
$\bullet$ Optimization needs to use subgradient techniques as the loss is not \\ fully differentiable \end{tabular}
& \begin{tabular}[c]{@{}>{\cellcolor[rgb]{0.929,0.867,0.929}}l@{}}
$\bullet$ Imbalanced dataset \\
$\bullet$ Precise segmentation boundaries \end{tabular}
\\
\rowcolor[rgb]{0.929,0.867,0.929} {\cellcolor[rgb]{0.835,0.675,0.835}} & & & &
\\
\rowcolor[rgb]{0.929,0.867,0.929} {\cellcolor[rgb]{0.835,0.675,0.835}}  
& Tversky Loss   
& \begin{tabular}[c]{@{}>{\cellcolor[rgb]{0.929,0.867,0.929}}l@{}}
$\bullet$ Adaptable to scenarios where type I or type II error is more critical
\\$\bullet$ Inherits advantages from IOU Loss\end{tabular}
& \begin{tabular}[c]{@{}>{\cellcolor[rgb]{0.929,0.867,0.929}}l@{}}
$\bullet$ Additional hyperparameters need fine-tuning for optimal performance
\\$\bullet$ Inherits gradient problems from IOU loss\end{tabular}   
& $\bullet$ Imbalanced error importance         
\\
\rowcolor[rgb]{0.929,0.867,0.929} {\cellcolor[rgb]{0.835,0.675,0.835}} & & & &
\\
\rowcolor[rgb]{0.929,0.867,0.929} {\cellcolor[rgb]{0.835,0.675,0.835}} 
& Focal Tversky Loss       
& \begin{tabular}[c]{@{}>{\cellcolor[rgb]{0.929,0.867,0.929}}l@{}}
$\bullet$ Combination of Focal loss and Tversky loss advantages
\\$\bullet$ Allows handling of class imbalances and weighting of FP and FN\end{tabular} 
& \begin{tabular}[c]{@{}>{\cellcolor[rgb]{0.929,0.867,0.929}}l@{}}
$\bullet$ Additional hyperparameter $\gamma$ requiring finetuning
\\$\bullet$ Close to convergence loss becomes suppressed and prevents model \\ from full convergence\end{tabular} 
& \begin{tabular}[c]{@{}>{\cellcolor[rgb]{0.929,0.867,0.929}}l@{}}
$\bullet$ Imbalanced error importance
\\$\bullet$ Imbalanced class distribution\end{tabular}    
\\
\rowcolor[rgb]{0.929,0.867,0.929} {\cellcolor[rgb]{0.835,0.675,0.835}} & & & &
\\
\rowcolor[rgb]{0.929,0.867,0.929} {\cellcolor[rgb]{0.835,0.675,0.835}}  
& Sensitivity Specificity Loss    
& $\bullet$ Able to control the tradeoff between FN and FP in imbalanced data    
& $\bullet$ Hyperparameter tuning of tradeoff parameter necessary   
& $\bullet$ Imbalanced data                                          
\\
\rowcolor[rgb]{0.929,0.867,0.929} {\cellcolor[rgb]{0.835,0.675,0.835}} & & & &
\\

\rowcolor[rgb]{0.929,0.867,0.929} {\cellcolor[rgb]{0.835,0.675,0.835}}  
& Region Mutual Loss      
& \begin{tabular}[c]{@{}>{\cellcolor[rgb]{0.929,0.867,0.929}}l@{}}
$\bullet$ Including relationships between pixels into the loss function
\\$\bullet$ Downsampling strategy to remain computational feasible\end{tabular}    
& $\bullet$ Computational overhead compared to simpler losses                     
& $\bullet$ General semantic segmentation                                          
\\
\rowcolor[rgb]{0.929,0.867,0.929} {\cellcolor[rgb]{0.835,0.675,0.835}} & & & &
\\
\rowcolor[rgb]{0.929,0.867,0.929}
\multirow{-28}{*}{{\cellcolor[rgb]{0.835,0.675,0.835}}\textbf{Region Level}} 
& Robust T-Loss
& \begin{tabular}[c]{@{}>{\cellcolor[rgb]{0.929,0.867,0.929}}l@{}}
$\bullet$ More resistance to noisy labels and outliers
\\$\bullet$ Learning of optimal tolerance level for label noise during training \end{tabular}
& $\bullet$ Simple alternative with worse performance
& \begin{tabular}[c]{@{}>{\cellcolor[rgb]{0.929,0.867,0.929}}l@{}}
$\bullet$ Noisy data
\\ $\bullet$ Data with outliers \end{tabular}
\\
\hline


\rowcolor[rgb]{1,0.957,0.773} {\cellcolor[rgb]{1,0.91,0.553}}
& Boundary Loss
& \begin{tabular}[c]{@{}>{\cellcolor[rgb]{1,0.957,0.773}}l@{}}
 $\bullet$ Simple combination with regional losses due to formulation \\ with regional softmax probabilities
 \\$\bullet$  Spatial regulation effect leads to smoother segmentation boundaries \end{tabular}
& $\bullet$ Limited to binary segmentation problems
& \begin{tabular}[c]{@{}>{\cellcolor[rgb]{1,0.957,0.773}}l@{}}
$\bullet$ Different-sized foreground and background
\\$\bullet$  Highly imbalanced class distributions \end{tabular}
\\

\rowcolor[rgb]{1,0.957,0.773} {\cellcolor[rgb]{1,0.91,0.553}}     
& Hausdorff Distance Loss             
& \begin{tabular}[c]{@{}>{\cellcolor[rgb]{1,0.957,0.773}}l@{}}
$\bullet$ Directly connected to common performance measure
\\$\bullet$ Variants exist which mitigates drawbacks\end{tabular}     
& \begin{tabular}[c]{@{}>{\cellcolor[rgb]{1,0.957,0.773}}l@{}}
$\bullet$ Sensitive to noise and outliers\\
$\bullet$ Focus only on minimizing the largest segmentation error degraded \\ general performance
\\$\bullet$ Difficulties with compound shapes\end{tabular}     
& \begin{tabular}[c]{@{}>{\cellcolor[rgb]{1,0.957,0.773}}l@{}}
$\bullet$ General semantic segmentation
\\$\bullet$ Medical image segmentation\end{tabular}     
\\
\rowcolor[rgb]{1,0.957,0.773} {\cellcolor[rgb]{1,0.91,0.553}}  & & & &
\\

\rowcolor[rgb]{1,0.957,0.773} {\cellcolor[rgb]{1,0.91,0.553}}   
& Boundary-aware Loss
& \begin{tabular}[c]{@{}>{\cellcolor[rgb]{1,0.957,0.773}}l@{}}
$\bullet$ Larger gradients around objects boundaries leads model \\ to pay attention to boundary regions \end{tabular}
& \begin{tabular}[c]{@{}>{\cellcolor[rgb]{1,0.957,0.773}}l@{}}
$\bullet$ Dependent on ground truth accuracy
\\$\bullet$ Reduces model robustness because of potential overfitting to \\ fine-grained boundaries
\\$\bullet$ Precise boundary localization required\end{tabular}           
& $\bullet$ Precise boundaries                            
\\
\rowcolor[rgb]{1,0.957,0.773} {\cellcolor[rgb]{1,0.91,0.553}}  & & & &
\\
\rowcolor[rgb]{1,0.957,0.773} {\cellcolor[rgb]{1,0.91,0.553}}      
& Active Boundary Loss      
& $\bullet$ Propagation of boundary information via distance transform improves boundary details
& $\bullet$ Boundary pixel conflicts degrade performance                       
& $\bullet$ Focus on the boundary details       
\\
\rowcolor[rgb]{1,0.957,0.773} {\cellcolor[rgb]{1,0.91,0.553}}  & & & &
\\
\rowcolor[rgb]{1,0.957,0.773} {\cellcolor[rgb]{1,0.91,0.553}}          
& InverseForm Loss
& \begin{tabular}[c]{@{}>{\cellcolor[rgb]{1,0.957,0.773}}l@{}}
 $\bullet$ Complements cross-entropy loss with boundary transformation
\\ $\bullet$ No increase in inference time compared to the cross-entropy loss
\\ $\bullet$ Distance-based measure outperform cross-entropy measures \end{tabular}
& \begin{tabular}[c]{@{}>{\cellcolor[rgb]{1,0.957,0.773}}l@{}}
 $\bullet$ Formulation can lead to exploding gradients limiting \\ search space for hyperparameters \end{tabular}    
& $\bullet$ General semantic segmentation
\\
\rowcolor[rgb]{1,0.957,0.773} {\cellcolor[rgb]{1,0.91,0.553}}  & & & &
\\
\rowcolor[rgb]{1,0.957,0.773} {\cellcolor[rgb]{1,0.91,0.553}}      
& Conditional Boundary Loss           
& $\bullet$ Enhances intra-class consistency and inter-class difference
& $\bullet$ Failure cases in mostly wrongly classified pixel regions
& \begin{tabular}[c]{@{}>{\cellcolor[rgb]{1,0.957,0.773}}l@{}}
 $\bullet$ General semantic segmentation
\\ $\bullet$ Noisy data \end{tabular}
\\
\rowcolor[rgb]{1,0.957,0.773} {\cellcolor[rgb]{1,0.91,0.553}}  & & & &
\\

\rowcolor[rgb]{1,0.957,0.773} {\cellcolor[rgb]{1,0.91,0.553}}   
& Boundary Difference Over Union Loss
& $\bullet$ Higher importance of boundary pixels leads to sharper boundaries
& $\bullet$ Hyperparameter requires different settings for smaller and larger objects
& $\bullet$ Data with large objects    
\\
\rowcolor[rgb]{1,0.957,0.773} {\cellcolor[rgb]{1,0.91,0.553}}  & & & &
\\
\rowcolor[rgb]{1,0.957,0.773} \multirow{-18}{*}{{\cellcolor[rgb]{1,0.91,0.553}}\textbf{Boundary Level}}   
& Region-wise Loss                      
& \begin{tabular}[c]{@{}>{\cellcolor[rgb]{1,0.957,0.773}}l@{}}
 $\bullet$ Flexible framework to assign importance values to different pixels
\\ $\bullet$ Advantages dependent on the concrete choice of region-wise maps \end{tabular}
& $\bullet$ Disadvantages dependent on the concrete choice of region-wise maps
& $\bullet$ Useful for many cases dependent
\\

\hline


\rowcolor[rgb]{0.804,0.922,0.969} {\cellcolor[rgb]{0.529,0.808,0.922}}   
& Combo Loss                        
& $\bullet$ Inherits advantages from Cross-Entropy Loss and Dice loss        
& $\bullet$ Hyperparameter tuning between losses and FP/FN importance
& \begin{tabular}[c]{@{}>{\cellcolor[rgb]{0.804,0.922,0.969}}l@{}}
$\bullet$ General semantic segmentation
\\$\bullet$ Lightly class imbalanced data\end{tabular}   
\\
\rowcolor[rgb]{0.804,0.922,0.969} {\cellcolor[rgb]{0.529,0.808,0.922}}   & & & &
\\

\rowcolor[rgb]{0.804,0.922,0.969} {\cellcolor[rgb]{0.529,0.808,0.922}} 
& Exponential Logarithmic Loss     
& \begin{tabular}[c]{@{}>{\cellcolor[rgb]{0.804,0.922,0.969}}l@{}}
$\bullet$ Logarithm and exponential of loss function allows control over pixel weights
\\$\bullet$ Focuses on less accurately predicted cases
\\$\bullet$ Inherits advantages from Cross-Entropy Loss and Dice loss\end{tabular}     
& $\bullet$ Suitable weighting of both loss terms necessary for optimal performance
& \begin{tabular}[c]{@{}>{\cellcolor[rgb]{0.804,0.922,0.969}}l@{}}
$\bullet$ General semantic segmentation
\\$\bullet$ Imbalanced class distributions
\\$\bullet$ Small or rare objects\end{tabular}    
\\
\rowcolor[rgb]{0.804,0.922,0.969} {\cellcolor[rgb]{0.529,0.808,0.922}}   & & & &
\\

\rowcolor[rgb]{0.804,0.922,0.969}
\multirow{-7}{*}{{\cellcolor[rgb]{0.529,0.808,0.922}}\textbf{Combination}}  
& Unified Focal Loss
& \begin{tabular}[c]{@{}>{\cellcolor[rgb]{0.804,0.922,0.969}}l@{}}
$\bullet$ Unifies hyperparameters to increase simplicity
\\ $\bullet$ Mitigates loss suppression and over-enhancement problem \end{tabular}
& \begin{tabular}[c]{@{}>{\cellcolor[rgb]{0.804,0.922,0.969}}l@{}}
 $\bullet$ Still sensitive to hyperparameter choice
\\ $\bullet$ Longer training time due to higher complexity \end{tabular}
& $\bullet$ Imbalanced class distributions                 
\\
\end{tabular}
}
\end{table*}

To investigate the use of loss functions for semantic segmentation, we look at the best-performing methods for both natural and medical image segmentation. For example,~\cite{wang2023internimage, yuan2020object, cheng2020panoptic, zhang2021dcnas,le2021regularized} rely on the Cross-Entropy loss function for optimization. \cite{chen2022vision} uses a Combo Loss with Cross-Entropy and Dice Loss,~\cite{li2019global} applies a TopK Loss. A custom hierarchical loss design is used by~\cite{borse2021hs3}. \cite{Borse2021InverseFormAL} applies their proposed InverseForm Loss and~\cite{mohan2021efficientps} a weighted per pixel loss introduced by~\cite{rota2017loss}. In medical image segmentation, most of the top performing models rely on a Combo Loss of Dice Loss and Cross-Entropy Loss~\cite{isensee2018nnu, roy2023mednext, liu2022phtrans, zhou2021nnformer, rahman2023multi, rahman2023medical, huang2021missformer}. \cite{irshad2023improved} uses a Combo Loss with Dice Loss and a Boundary Loss. A combination of L1 loss, cross-entropy loss, and a 3D contrastive coding loss is used for pretraining by~\cite{tang2022self}. \cite{zhou2019prior} uses a Prior-Aware Loss that measures the matching probability of two distributions via the Kullback-Leibler divergence. We see that the applied loss functions change depending on the task, while in urban street scenes, the simple Cross-Entropy loss function is dominant, and in organ segmentation, the Dice Loss is additionally used since the Dice Score is the common evaluation metric.

\Cref{acdc} presents visual representations that illustrate the qualitative results of Boundary Difference over Union (Boundary DoU) loss and several other loss functions. The figure shows the clear advantage of using appropriate loss functions for segmenting complicated regions. Specifically, we can observe a more accurate localization and segmentation for boundary regions in boundary-level loss functions.  In addition, the significant shape variations of the right ventricle (RV) region in Rows 1, 3, 4, and 6 can lead to problems of both under- and mis-segmentation. In this case, the boundary DoU loss function effectively addresses this challenge compared to alternative loss functions. Conversely, the and myocardium (MYO) region has an annular structure with highly detailed regions, as seen in Rows 2 and 5. In these cases, other loss functions tend to produce varying degrees of under-segmentation, while the boundary DoU loss function provides a more comprehensive segmentation. The reduction in misclassification and underclassification ultimately increases the potential for improved clinical guidance.

\begin{figure*}[h]
    \centering
    \includegraphics[width=\textwidth]{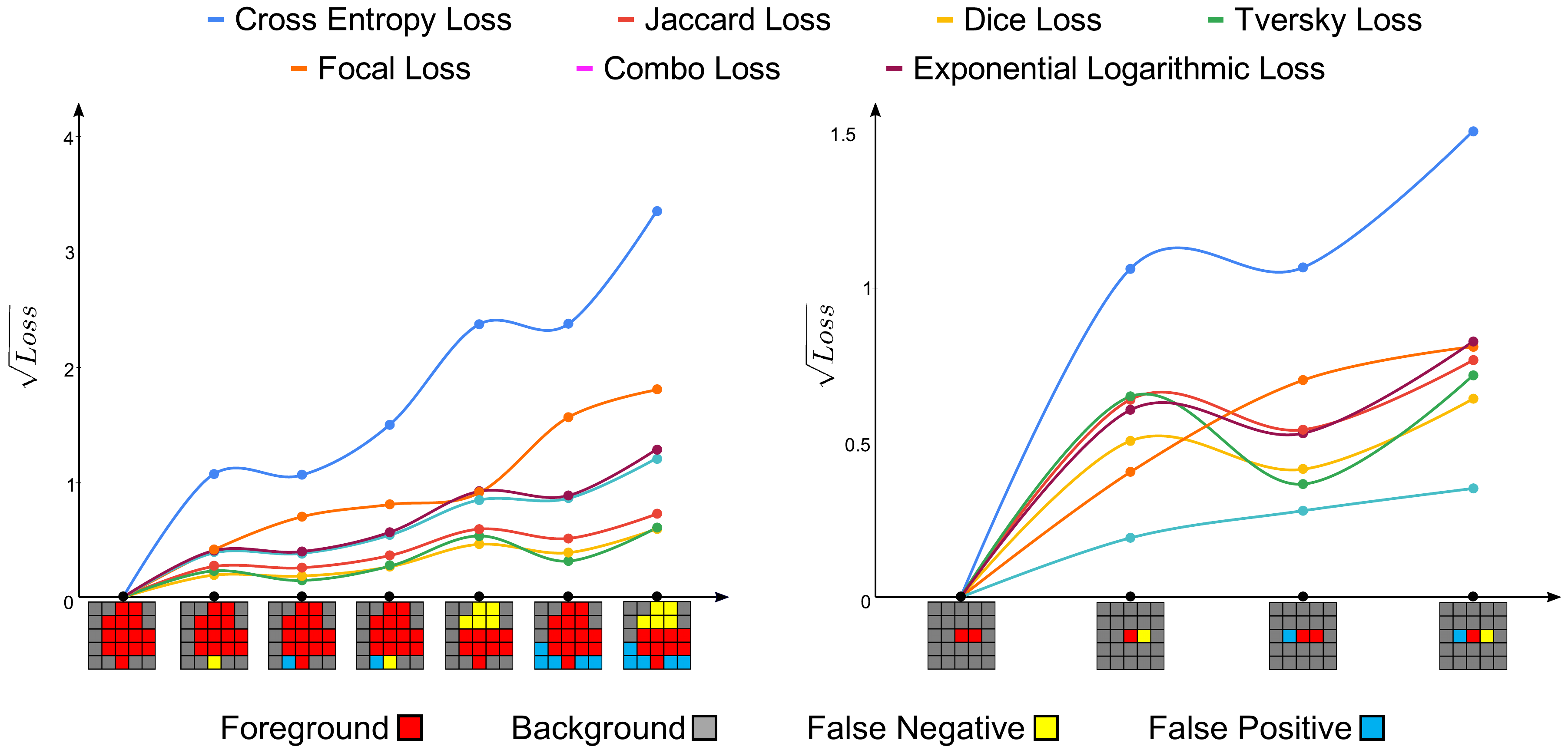}
    \caption{Impact of different loss functions on the segmentation of large (left plot) and small (right plot) objects.~\cite{asgari2021deep}.}
    \label{fig:losscompare}
\end{figure*}

To further explore the impact of different loss functions on segmentation performance, we turn our attention to \Cref{fig:losscompare}, which provides a visual representation of how different loss functions perform in the segmentation of both large and small objects~\cite{asgari2021deep}. As we move from left to right in the plots, we can see a gradual decrease in the overlap between the predictions and the ground truth mask. This decrease leads to the appearance of more false positives and false negatives. Ideally, the loss values should show a consistent upward trend as the number of false positives and negatives increases.

For large objects, most of the employed loss functions used adhere to this ideal scenario. However, for small objects (as shown in the right plot), only the combo loss and the Focal loss exhibit a more pronounced penalization of larger errors in a monotonic fashion. In simpler terms, the functions based on overlap measures show considerable variation when used to segment both small and large objects. These results underscore the critical notion that the choice of a loss function depends on the size of the object of interest~\cite{asgari2021deep}.

This observation prompts us to consider how these findings can guide practitioners in selecting the most appropriate loss function based on the specific characteristics and sizes of the objects they wish to segment, thus shedding further light on the nuanced relationship between loss functions and segmentation performance.

Overall, both \Cref{fig:losscompare} and \Cref{acdc} clarify the paramount importance of loss function choice in obtaining a more stable segmentation on the hard-to-segment objects, verifying the capabilities and distinct applications of each of the previously mentioned losses within their respective domains.

In addition to the discussed loss functions, further advancements in model performance can be achieved by integrating supplementary loss functions tailored for specific tasks or by adapting the existing ones to the task at hand. For example, in \cite{demir2023topology}, the author introduces a novel loss function called Topology-Aware Focal Loss (TAFL), which combines the conventional Focal Loss with a topological constraint term based on the Wasserstein distance between ground truth and predicted segmentation masks' persistence diagrams. This incorporation ensures the preservation of identical topology as the ground truth, effectively addressing topological errors, while simultaneously handling class imbalance. Another approach, as demonstrated by Wen et al in~\cite{wen2022pixel}, proposes a straightforward yet effective method named pixel-wise triplet learning. This method focuses on improving boundary discrimination without introducing additional computational complexity. By employing pixel-level triplet loss, segmentation models can learn more discriminative feature representations at boundaries. Notably, this method can be seamlessly integrated into state-of-the-art segmentation networks, serving as a versatile boundary booster applicable to both binary and multiclass medical segmentation tasks.
Ultimately, the selection of a loss function for semantic segmentation tasks can be tailored to the learning algorithm employed. For instance, in the context of recent diffusion-based generative models \cite{kazerouni2023diffusion}, leveraging a more sophisticated loss function can yield improvements not only in segmentation performance but also in enhancing the reconstruction process. Likewise, in implicit neural representation \cite{molaei2023implicit}, adapting the loss function can contribute to an efficient segmentation task. 

\section{Experiments}
\label{sec:experiment}
\subsection{Experimental Setup}
We train the models on a single RTX 3090 GPU using the Pytorch library. Stochastic gradient descent is applied with a batch size of 8 and a base learning rate of 0.01 for a total of 300 epochs. Furthermore, we use deterministic training with a fixed seed setting to obtain comparable results and to avoid other variations caused by randomness. The networks are trained using a combination of cross-entropy loss $L_{CE}$ and varying loss function $L_{var}$ resulting in the total loss of:
\begin{equation}
    L_{total} = 0.5 * L_{var} + 0.5 * L_{CE}.
\end{equation}
We evaluate the training convergence and performance of 6 different loss functions, i.e. Dice loss, Focal loss, Tversky loss, Focal Tversky loss, Jaccard loss and Lovász-softmax loss.

\subsection{Datasets and Evaluation Metrics}
For comparison, we use two common publicly available datasets, Synapse and Cityscapes. The former is a medical image segmentation dataset consisting of 30 abdominal CT scans. The performance is evaluated by using the Dice Score Coefficient (DSC) and Hausdorff Distance (HD) metrics, considering 8 abdominal organs, i.e., aorta, gallbladder, left and right kidney, liver, pancreas, spleen and stomach. The latter is an urban street segmentation dataset with 30 classes and 5000 fine annotated images, where the mean Intersection over Union (mIoU) is used as the evaluation metric. 

\subsection{Experimental Results}
We perform the evaluation on two common deep learning models, i.e. a traditional one with UNet~\cite{ronneberger2015unet} and a model based on the Vision Transformer architecture with TransUNet~\cite{chen2021transunet}.

\subsubsection{Quantitative Results}
The detailed performance results are shown in Table \ref{table:LossSynapseOverview}.
\begin{table*}[!ht]
    \begin{center}
    \caption{Comparison results of the different loss functions evaluated on the Synapse dataset. The best results across the hyperparameter variations are reported. \textcolor{blue}{Blue} indicates the best result, and \textcolor{red}{red} displays the second-best. DSC are presented for abdominal organs spleen (Spl), right kidney (RKid), left kidney (LKid), gallbladder (Gal), liver (Liv), stomach (Sto), aorta (Aor) and pancreas (Pan).}
        \resizebox{\textwidth}{!}{
        \begin{tabular}{l|c c c c c c c c|cc}
        \toprule
            \multirow{2}{*}{Loss Function} & \multirow{2}{*}{Aor} &  \multirow{2}{*}{Gal} &  \multirow{2}{*}{LKid} & \multirow{2}{*}{RKid}  & \multirow{2}{*}{Liv}  & \multirow{2}{*}{Pan} & \multirow{2}{*}{Spl} &  \multirow{2}{*}{Sto} &  \multicolumn{2}{c}{Average} 
            
            \\ \cmidrule{10-11}
             & & & & & & & & & DSC $\uparrow$ & HD95 $\downarrow$ \\
                \midrule
                \midrule
        & \multicolumn{8}{c}{\textbf{UNet}} & & \\
        \midrule
        Dice Loss~ & 85.46 & 59.39  & 81.23 & 73.53 & 93.26 & 48.59  & 84.13 & 69.14 & 74.34 & 37.68 \\
        Focal Loss~ & 85.12 & 59.82 & \textcolor{blue}{82.10} & 74.74 & 93.64 & 47.01 & 83.56 & 70.72  & 74.59 & 35.10 \\
        Tversky Loss~ & 88.57 & \textcolor{red}{70.22} & 79.39 & \textcolor{blue}{78.38} & \textcolor{red}{94.41} & \textcolor{blue}{69.45} & \textcolor{red}{88.50} & 76.71 & \textcolor{blue}{80.70} & \textcolor{blue}{27.16}\\
        Focal Tversky Loss~ & 88.51 & 69.90 & 79.44 & 74.30 & 93.69 & \textcolor{red}{66.04} & 86.86 & \textcolor{red}{76.80} & 79.44 & 30.97 \\
        Jaccard Loss~ & \textcolor{blue}{89.08} & 67.69 & \textcolor{red}{81.60} & \textcolor{red}{77.98} & \textcolor{blue}{95.04} & 65.45 & \textcolor{blue}{88.88} & \textcolor{blue}{78.32} & \textcolor{red}{80.52} & \textcolor{red}{27.39} \\
        Lovász-Softmax Loss~& \textcolor{red}{88.87} & \textcolor{blue}{72.89} & 81.01 & 73.99 & 94.38 & 65.36 & 86.78 & 75.74 & 79.88 & 31.43 \\
        \midrule
        \midrule
        & \multicolumn{8}{c}{\textbf{TransUNet}} & & \\

        \midrule
        Dice Loss~ & 84.99 & 53.10 & 78.89 & 75.62 & 93.08 & 48.26 & 85.55 & 74.39 & 74.23 & 41.64 \\
        Focal Loss~ & 86.96 & 53.42 & 76.48 & 73.77 & 91.29 & 49.18 & 84.42 & 74.77 & 73.79 & 45.66 \\
        Tversky Loss~ & 86.40 & \textcolor{red}{66.44} & \textcolor{red}{83.41} & \textcolor{red}{79.07} & \textcolor{red}{93.95} & \textcolor{blue}{64.03} &  \textcolor{red}{89.21} &  \textcolor{blue}{80.04} & \textcolor{red}{80.32} & 31.76\\
        Focal Tversky Loss~&  \textcolor{red}{87.89} & 66.22 & 81.58 & 76.73 & 93.41 & \textcolor{red}{63.63} & 89.06 & 77.62 & 79.52 & 30.48 \\
        Jaccard Loss~ &  \textcolor{blue}{88.63} &  \textcolor{blue}{67.52} &  \textcolor{blue}{83.42} &  \textcolor{blue}{80.88} &  \textcolor{blue}{94.42} & 62.53 & \textcolor{blue}{90.59} & \textcolor{red}{78.83} &  \textcolor{blue}{80.85} &  \textcolor{blue}{27.47} \\
        Lovász-Softmax Loss~& 87.18 & 64.34 & 76.89 & 76.06 & 93.31 & 59.29 & 88.96 & 73.56 & 77.45 & \textcolor{red}{29.84} \\
        \midrule
        \bottomrule
        \end{tabular}
        }\vspace{0.5em}
        
        \label{table:LossSynapseOverview}
        \end{center}
\vspace{-0.8cm}
\end{table*}

We observe that for both the UNet and TransUNet there is a significant gap in performance between the different loss functions. For UNet, this gap is at a maximum of 6.36\% DSC, with the Tversky Loss performing best and the Dice Loss performing worst. On the TransUNet side, this difference is 7.06\% DSC points, wiht Jaccard Loss performing best and the Focal Loss performing worst. Furthermore, the Dice Loss and the Focal Loss have reduced performance for small organs such as the pancreas and the gallbladder, while the other losses perform significantly better. This highlights the advantages of the Jaccard and Tversky Losses, which are able to produce sharper segmentation boundaries due to their overlapping nature and direct relationship to the segmentation performance. Other losses, such as Focal Loss, cannot benefit from their characteristics because the class imbalance of the abdominal scans is generally low, resulting in lower performance. We do not observe a noticeable difference between the two trained networks, which leads to the conclusion that the choice of loss is more data-dependent than network-dependent.

\begin{figure*}[!ht]
     \centering
     \includegraphics[width=0.49\linewidth, trim=55px 15px 50px 50px, clip]{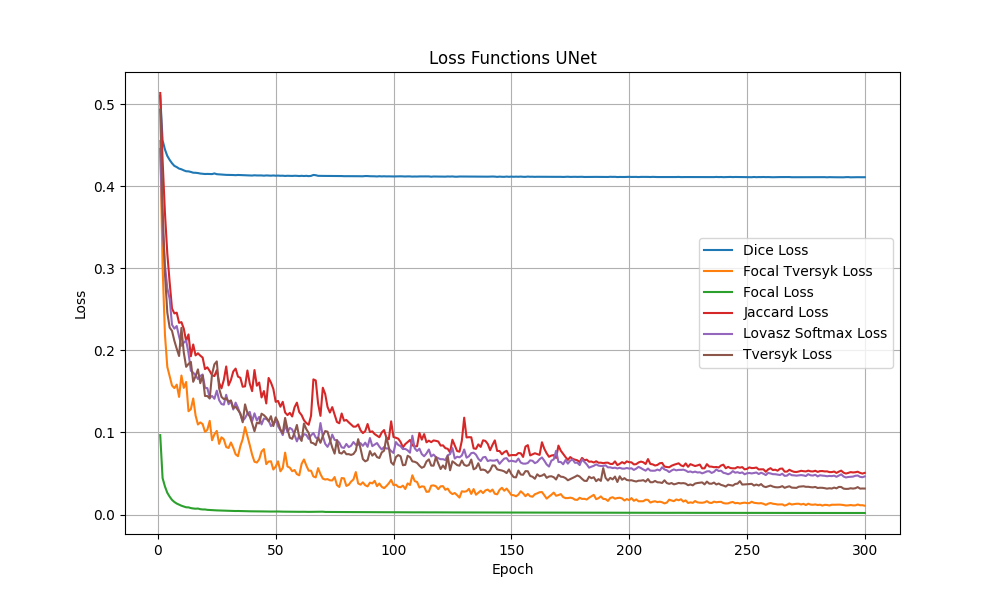}
     \includegraphics[width=0.49\linewidth, trim=55px 15px 50px 50px, clip]{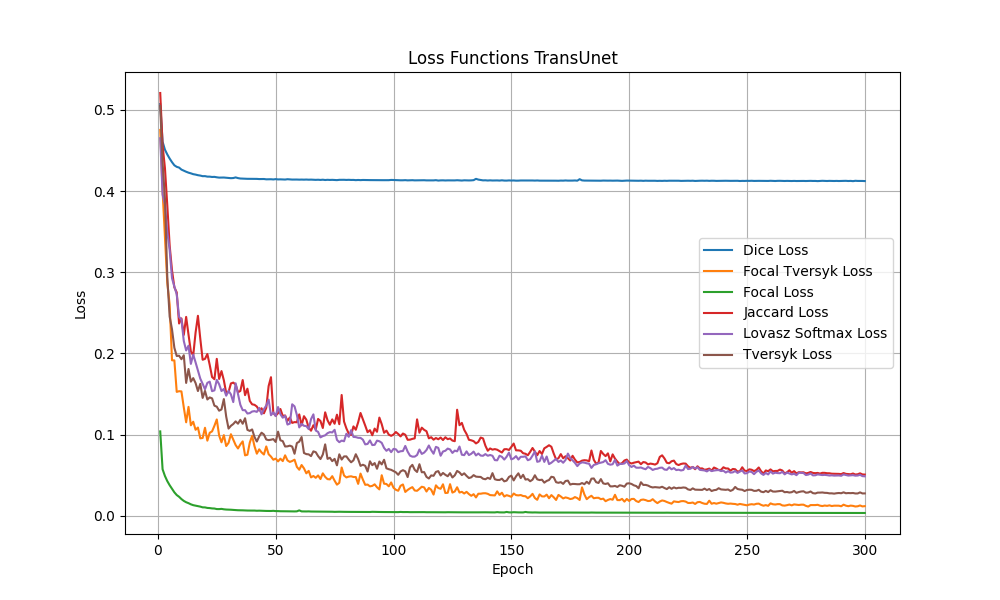}
     \caption{Loss function plots trained on the Synapse dataset for the UNet network on the left and TransUNet network on the right for the selected loss functions, i.e. Dice Loss, Focal Tversky Loss, Focal Loss, Jaccard Loss Lovász-Softmax Loss and Tversky. The loss values are averaged over all iterations in one epoch.}
     \label{fig:lossvalues_UNet_TransUNet}
\end{figure*}

Moreover, we examine the loss behavior during training and show the loss values overall training epochs in Figure~\ref{fig:lossvalues_UNet_TransUNet}.
Dice Loss and Focal Loss show constant behavior, which explains their overall worse performance, as the model training gets stuck early on in the training process. The other four losses show typical training behavior, with training converging around 200 epochs. Again, there is no significant difference between the UNet and TransUNet models.

The Cityscapes performance results are shown in Table \ref{table:LossCityscapesOverview}.

\begin{table}[!ht]
    \centering
    \caption{Comparison results of the different loss functions evaluated on the Cityscapes dataset. \textcolor{blue}{Blue} indicates the best result, and \textcolor{red}{red} displays the second-best.}
    
    \resizebox{0.45\textwidth}{!}{%
        \begin{tabular}{l|cc|l|cc}
            \toprule
            \textbf{Model} & \multicolumn{2}{c|}{\textbf{UNet}} & \textbf{} & \multicolumn{2}{c}{\textbf{TransUNet}} \\
            \midrule
            \textbf{Loss Function} & \textbf{mIoU} & & & \textbf{mIoU} & \\
            \midrule
            \midrule
            Dice Loss & \textcolor{red}{63.28} & & & 56.42 & \\
            Focal Loss & 63.03 & & & 56.59 & \\
            Tversky Loss & \textcolor{blue}{63.52} & & & \textcolor{red}{58.58} & \\
            Focal Tversky Loss & 62.52 & & & 58.48 & \\
            Jaccard Loss & 63.01 & & & \textcolor{blue}{58.97} & \\
            Lovász-Softmax Loss & 62.54 & & & 56.67 & \\
            \midrule
            \bottomrule
        \end{tabular}%
    }
    \label{table:LossCityscapesOverview}
\end{table}

In the UNet case, Tversky Loss and Dice Loss are the best performing loss functions while the discrepancy between all losses is 1\% DSC. For the TransUNet case, Jaccard Loss is the best performing and Tversky the second best, and the discrepancy is 2.55\% DSC. The difference in discrepancy between the models shows that the importance of the choice of loss function varies with the model chosen, but the TransUNet case shows that it is important to experiment with this choice to improve model performance.

\begin{figure*}[!ht]
     \centering
     \includegraphics[width=0.49\textwidth]{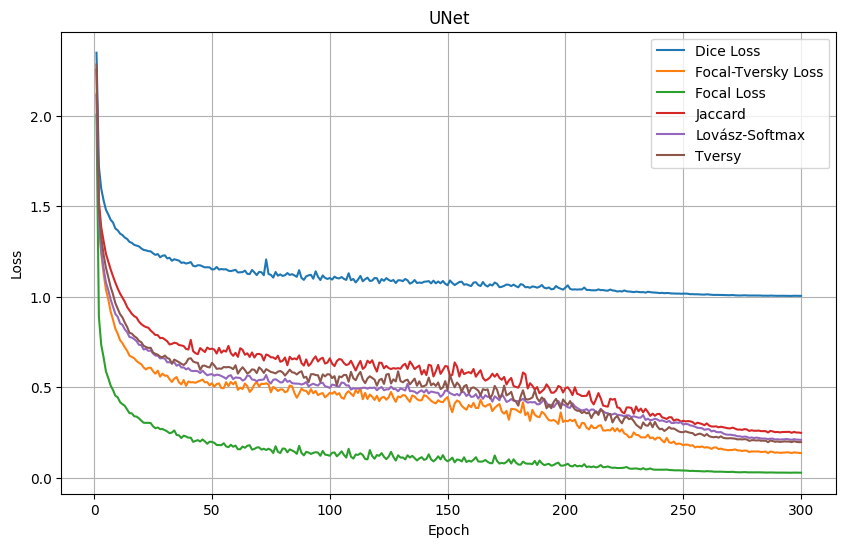}
     \includegraphics[width=0.49\textwidth]{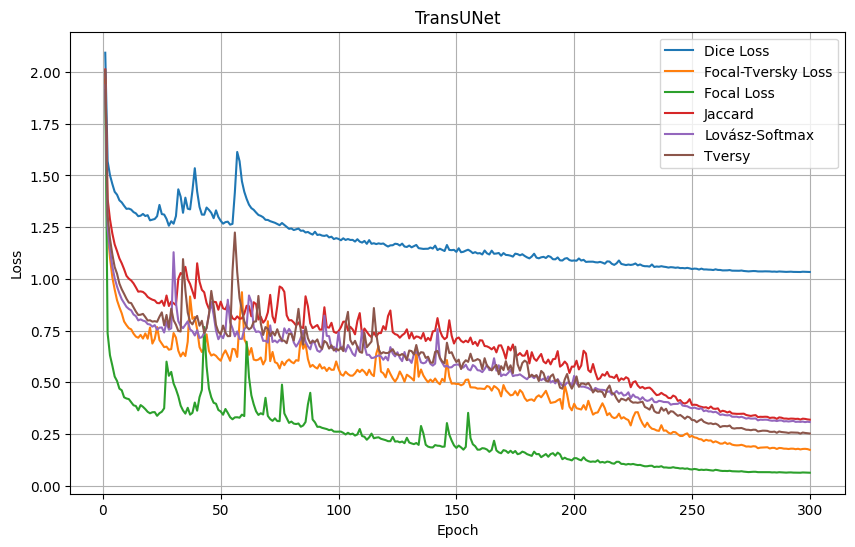}
     \caption{Loss function plots trained on the Cityscapes dataset for the UNet network on the left and TransUNet network on the right for the selected loss functions, i.e. Dice Loss, Focal Tversky Loss, Focal Loss, Jaccard Loss, Lovász-Softmax Loss and Tversky. The loss values are averaged over all iterations in one epoch.}
     \label{fig:lossvaluesCityscapes_UNet_TransUNet}
\end{figure*}

The loss behavior of these training runs is illustrated in \Cref{fig:lossvaluesCityscapes_UNet_TransUNet}.
It behaves differently compared to the Synapse evaluation. Dice Loss and Focal Loss decrease over the epochs, showing actual training progress instead of constant values. This explains the better performance on the Cityscapes dataset. Furthermore, at about 250 epochs of the TransUNet and UNet training, another decrease is observed for Jaccard, Lovász-Softmax, Tversky and Focal Tversky Loss, leading to significant later convergence. This underscores the fact that the loss performance is highly model-dependent and that some losses may be suitable choices for one network but are not suitable for others.

\subsubsection{Qualitative Results}
\begin{figure*}[!ht]
     \centering
     \includegraphics[width=\textwidth]{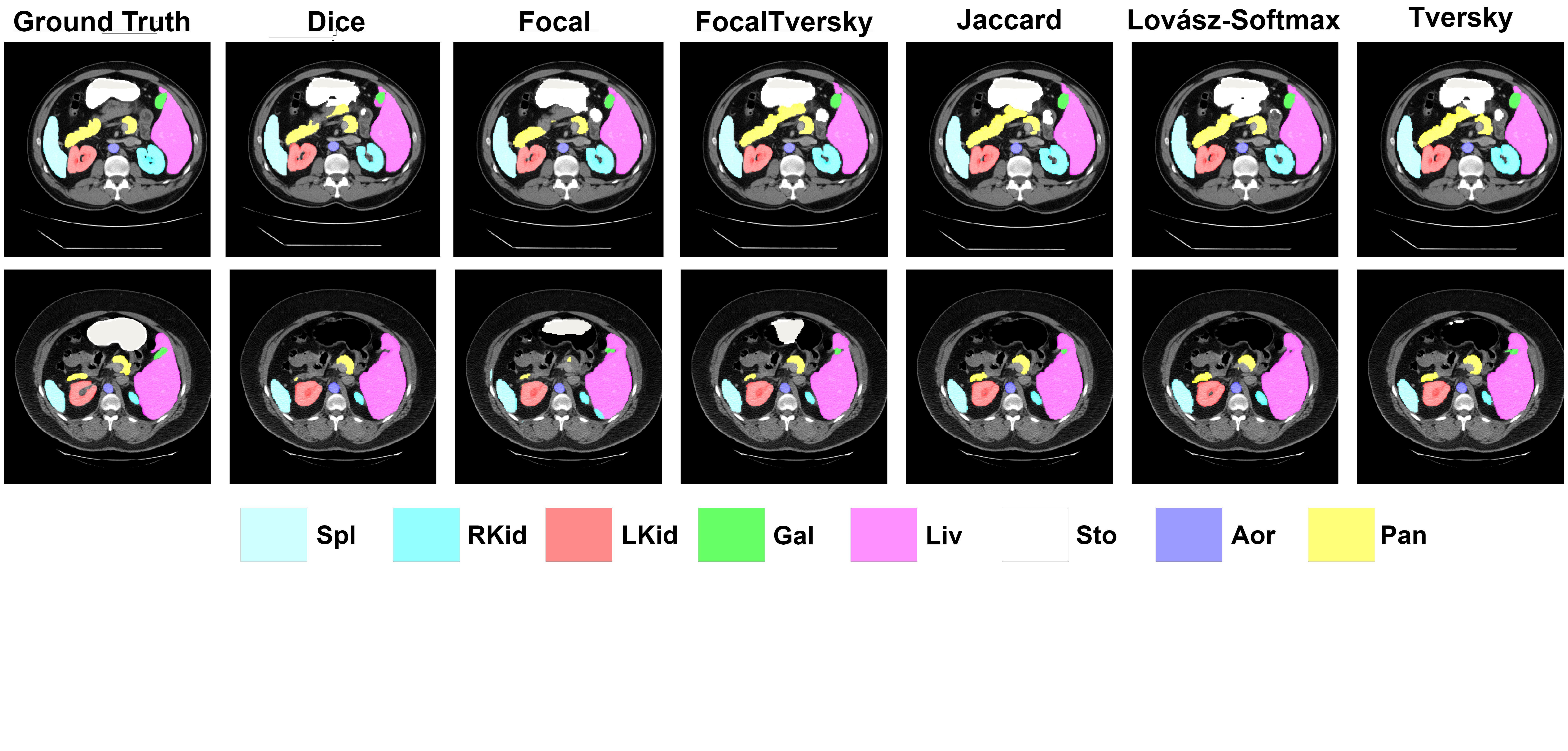}
     \caption{Qualitative visualization of the TransUNet network trained with various loss function on the Synapse dataset.}
     \label{fig:Synapse_qualitative_results}
\end{figure*}

The segmentation masks of the networks trained with different loss functions on the Synapse dataset are visualized in \Cref{fig:Synapse_qualitative_results}.
Compared to the ground truth segmentation, the Dice Loss shows varying performance, while in the top example, the segmentation appears quite good except for the pancreas region. In the lower example, it completely fails to identify the stomach and the gallbladder. 
The Focal Tversky Loss, on the other hand, presents the most promising segmentation map, correctly identifying all organs, with only minor variations observed in the stomach. Conversely, the Jaccard, Lovász-Softmax, and Tversky losses also struggle to segment this area. This empirical analysis leads to the conclusion that the Focal Tversky Loss is the most suitable choice. It excels in rectifying both misidentifications and omissions, thereby improving the segmentation of complex and misclassified cases.

\begin{figure*}[!ht]
    \centering
    \includegraphics[width=\textwidth]{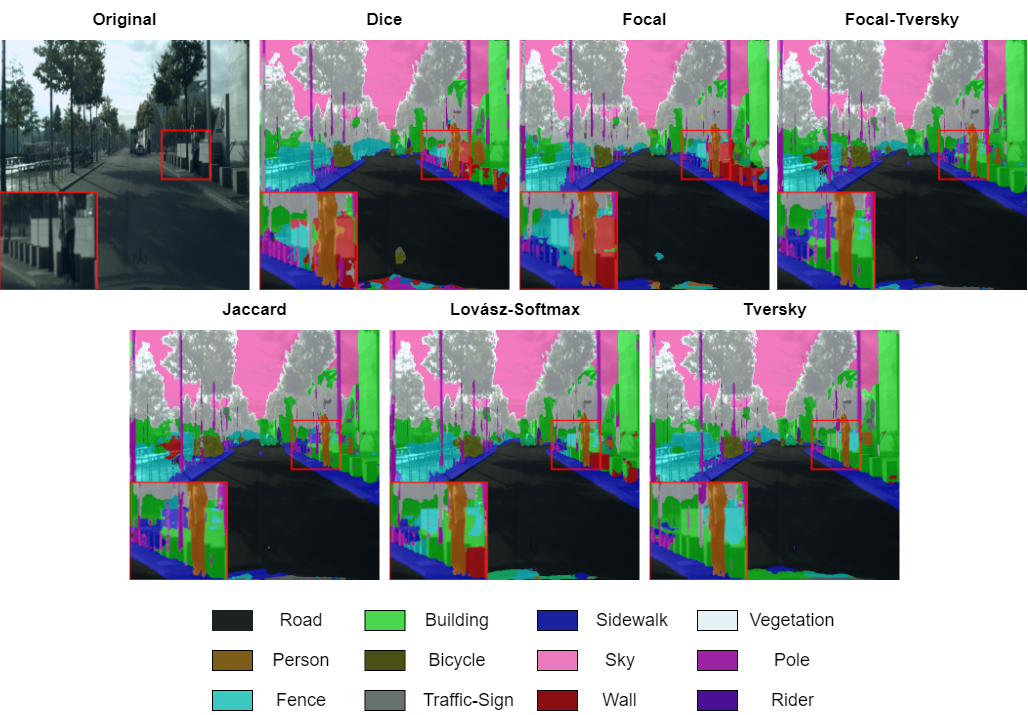}
    \caption{Qualitative visualization of the TransUNet network trained with various loss functions on the Cityscapes dataset. A region of interest is zoomed in for better visualization at the bottom left of each image.}
    \label{fig:Cityscapes_qualitative_results}
\end{figure*}

For the Cityscapes dataset, qualitative results are illustrated in \Cref{fig:Cityscapes_qualitative_results}.
Similar results can be observed in all the networks, where the sidewalk border posts are sometimes classified as buildings, walls or poles because there is no class of its own. Bicycle racks are also mostly classified as fences or walls because there is no more precise class. Overall, there are only minor misclassifications across all damages, and the segmentation boundaries are more or less accurate. This also reflects the quantitative results presented in the previous section.

\subsubsection{On the importance of hyperparameters}
We explore the importance of hyperparameter choice by performing final performance comparisons of different hyperparameter choices on the Synapse dataset. \Cref{tab:hyperparameter_variations}
shows the performance results for the Focal Loss, Tversky Loss and the Focal Tversky Loss. 

\begin{table}[!ht]
    \centering
    \caption{Results of the variation of the hyperparameter settings in the Focal Tversky Loss, Focal Loss, and Tversky Loss evaluated on TransUNet. \textcolor{blue}{Blue} indicates the best result, and \textcolor{red}{red} displays the second-best.}
    \resizebox{0.95\columnwidth}{!}{%
        \begin{tabular}{ccc|ccc|ccc}
            \toprule
            \multicolumn{3}{c|}{\textbf{Focal Tversky Loss}} & \multicolumn{3}{c|}{\textbf{Focal Loss}} & \multicolumn{3}{c}{\textbf{Tversky Loss}} \\
            \textbf{Parameter setting} & \textbf{DSC}$\uparrow$ & \textbf{HD95}$\downarrow$ & \textbf{Parameter setting} & \textbf{DSC}$\uparrow$ & \textbf{HD95}$\downarrow$ & \textbf{Parameter setting} & \textbf{DSC}$\uparrow$ & \textbf{HD95}$\downarrow$ \\
            \midrule
            \midrule
            $\alpha = 0.2$ / $\gamma = 2$ & 77.65 & 38.13 & $\gamma = 1.5$ & 73.27 & 48.20 & $\alpha = 0.2$ & \textcolor{red}{80.32} & 31.76 \\
            $\alpha = 0.4$ / $\gamma = 2$ & \textcolor{red}{78.89} & 31.53 & $\gamma = 2.0$ & \textcolor{blue}{73.79} & \textcolor{red}{45.66} & $\alpha = 0.4$ & \textcolor{blue}{80.53} & \textcolor{red}{25.94} \\
            $\alpha = 0.6$ / $\gamma = 2.0$ & \textcolor{blue}{79.52} & \textcolor{red}{30.48} & $\gamma = 2.5$ & 73.36 & 48.64 & $\alpha = 0.6$ & 79.57 & 31.40 \\
            $\alpha = 0.8$ / $\gamma = 2$ & 77.53 & \textcolor{blue}{29.36} & $\gamma = 3.0$ & \textcolor{red}{73.60} & \textcolor{blue}{41.60} & $\alpha = 0.8$ & 80.05 & \textcolor{blue}{24.09} \\
            \midrule
            \bottomrule
        \end{tabular}%
    }
    \label{tab:hyperparameter_variations}
\end{table}

\begin{figure*}[!ht]
    \centering
    \begin{subfigure}{0.51\textwidth}
        \centering
     \includegraphics[width=\linewidth, trim=45px 15px 50px 35px, clip]{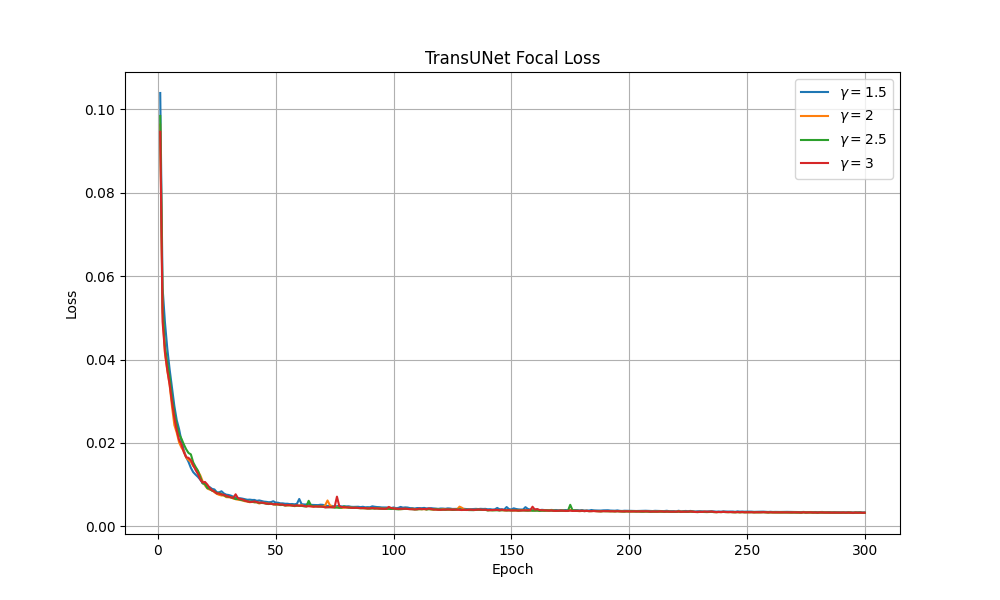}
        \caption{Focal Loss}
        \label{subfig:FocalLoss}
    \end{subfigure}%
    \begin{subfigure}{0.51\textwidth}
        \centering
        \includegraphics[width=\linewidth, trim=45px 15px 50px 35px, clip]{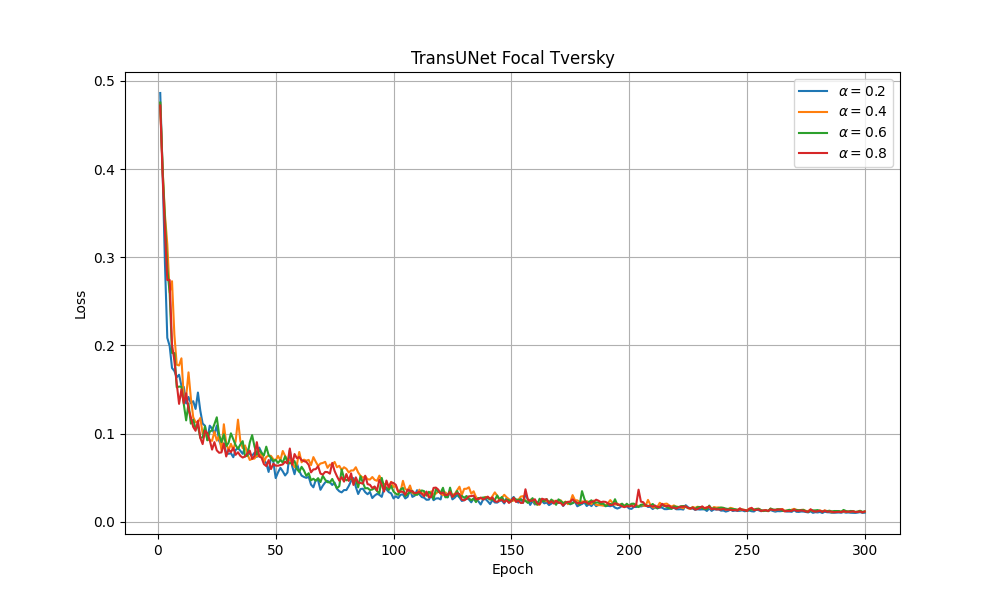}
        \caption{Focal Tversky Loss}
        \label{subfig:FocalTverskyLoss}
    \end{subfigure}
    
    \begin{subfigure}{\textwidth}
        \centering
        \includegraphics[width=0.7\linewidth, trim=45px 15px 50px 35px, clip]{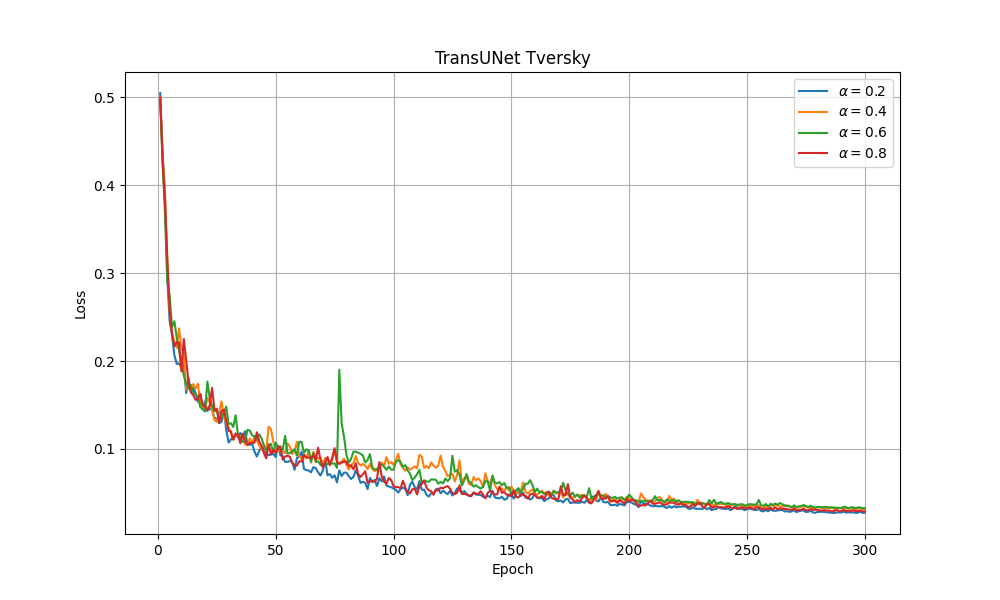}
        \caption{Tversky Loss}
        \label{subfig:TverskyLoss}
    \end{subfigure}
    
    \caption{Loss function plots for the hyperparameter variations for the Focal Loss, Focal Tversky Loss, and the Tversky Loss. The loss values are averaged over all iterations in one epoch.}
    \label{fig:Hyperparameter_variations}
\end{figure*}

These results show that the choice of hyperparameters leads to a maximum difference of 1.99\% DSC for the Focal Tversky Loss, 0.33\% DSC for the Focal Loss, and 0.96\% for the Tversky Loss, which shows varying importance. Looking more closely at the loss characteristics shown in Figure~\ref{fig:Hyperparameter_variations}, the generally constant and approximately equal behavior of the Focal Loss explains the small amount of variation in performance. The Focal Tversky and Tversky Loss show similar but slightly different training behavior, resulting in a larger performance gap.

Overall, this shows that hyperparameters need to be chosen carefully as they can significantly affect the resulting performance. To choose them optimally, a fine parameter sweep is necessary, as there are usually no general guidelines, which is a common drawback of loss functions that include parameter choices.

\section{Future Work and Open Challenges}
\label{sec:future}
Despite the developments in semantic segmentation, especially in the proposal of better loss functions, they are still limited in several aspects and require further research efforts to become viable for practical applications. In the following, we briefly discuss some limitations and future directions.

\subsection{Hyperparameter guidelines}
Since a large part of the losses require hyperparameter values and a suitable choice can significantly increase the model performance, further research in this area can assist developers in their model design process by providing guidelines or recommendations for hyperparameter choices. This requires extensive studies in different semantic segmentation domains and with different underlying network architectures to evaluate the impact of hyperparameters, their data and model dependencies.

\subsection{Combo Loss studies}
The number of loss functions based on a combination of other losses in academic literature is small. Since combo losses can inherit advantages from each of their underlying losses, while possibly mitigating disadvantages, they can be an appropriate choice for many models and tasks. Research exploring new combinations of recent loss functions can potentially create novel loss functions that improve performance.

\subsection{Interactivity with Label Uncertainty}
Many semantic segmentation applications involve dealing with ambiguous or uncertain labels. To address this, future research can explore loss functions that incorporate measures of label uncertainty or ambiguity. These novel loss functions could dynamically adapt to the reliability of ground truth annotations by assigning adaptive weights to each label based on its certainty. This adaptability is particularly important in scenarios where human annotators may provide labels with varying degrees of confidence, helping to mitigate the effects of noisy or uncertain data.

\subsection{Robustness to Noisy Annotations}
In practice, obtaining perfectly accurate annotations for training data is often challenging. Robust loss functions that are less sensitive to label noise or errors can be a game-changer. Research in this area could focus on developing loss functions that can automatically identify and down-weight noisy annotations during training. Additionally, exploring techniques that combine loss functions with data augmentation strategies to improve model resilience to noisy data can further improve segmentation performance.

\subsection{Adaptation to Pre-trained Foundation Models}
As foundation models, such as CLIP~\cite{radford2021learning}, Stable Diffusion~\cite{rombach2022high}, GPT~\cite{openai2023gpt4}, etc. are increasingly being used as off-the-shelf frameworks for various downstream tasks, it is important to conduct studies on how to adapt semantic segmentation losses to these pre-trained models. By exploring and analyzing the potential methods and approaches to fine-tune these generic models, we can ensure optimal performance and accuracy in various downstream medical applications.

\subsection{Losses and Metrics: Fairness}
The performance of a loss function is evaluated using one or multiple segmentation metrics. It is worth including more evaluation metrics for semantic segmentation tasks and further investigating which loss function would be more advantageous for each evaluation metric. Moreover, we can use all other segmentation loss functions to evaluate the one under investigation in a leave-one-out cross-validation (LOOCV) manner. This would allow for a comprehensive comparison and analysis of the performance of different loss functions, providing a more robust assessment of their effectiveness.

\section{Conclusion}
\label{sec:conclusion}

In conclusion, this survey provides a comprehensive overview of $25$ loss functions for semantic segmentation, with a focus on their applications in medical and natural images. We have emphasized the critical role these loss functions play in improving segmentation models. We have introduced a structured taxonomy, performed validation experiments on popular datasets, identified open challenges and areas for future research, and highlighted recent developments beyond $2020$. This survey serves as a valuable resource for researchers and practitioners, providing insights to guide loss function selection and further innovation in semantic segmentation.

\bibliographystyle{unsrt}
\bibliography{Ref}

\end{document}